\documentclass[10pt,twocolumn,letterpaper]{article}

\usepackage{iccv}       
\usepackage[accsupp]{axessibility}
\usepackage{mathtools}
\usepackage{amssymb}
\usepackage{enumitem}
\usepackage{amsfonts}
\usepackage{pifont}
\usepackage{amsopn}
\usepackage{graphicx}
\usepackage{textcomp}
\usepackage{xfrac}
\usepackage{ifthen}
\usepackage{xspace}
\usepackage{bbm}
\usepackage{wrapfig}
\usepackage[ruled,vlined,linesnumbered]{algorithm2e}
\usepackage{overpic}
\usepackage{currfile} %
\usepackage{colortbl}  
\currfiledir
\usepackage{multirow}
\SetAlFnt{\small}
\SetAlCapFnt{\small}
\SetAlCapNameFnt{\small}
\usepackage{booktabs}
\usepackage{footnote}
\usepackage{autobreak}
\usepackage[thicklines]{cancel}
\definecolor{green}{rgb}{0, 0.4, 0}
\definecolor{orange}{rgb}{0.8, 0.6, 0.2}
\definecolor{red}{rgb}{1.0, 0.0, 0.0}
\definecolor{teal}{rgb}{0.0, 0.4, 0.4}
\definecolor{purple}{rgb}{0.65,0,0.65}
\definecolor{saffron}{rgb}{0.95,0.75,0.2}
\definecolor{turquoise}{rgb}{0.0,0.5,0.5}
\definecolor{brown}{rgb}{0.5, 0.16, 0.16}

\newlength\savedwidth

\usepackage[normalem]{ulem}

\newcommand{\hidecomment}[1]{}

\newcommand{\cI}{\mathcal{I}}
\newcommand{\cC}{\mathcal{C}}

\newcommand{\cS}{\mathcal{S}}
\newcommand{\cM}{\mathcal{M}}
\newcommand{\cN}{\mathcal{N}}

\newcommand{\cE}{\mathcal{E}}
\newcommand{\cA}{\mathcal{A}}
\newcommand{\cL}{\mathcal{L}}

\newcommand{\cP}{\mathcal{P}}

\newcommand{\name}{CogNav}
\def\paperID{7286} 
\def\confName{ICCV}
\def\confYear{2025}

\title{CogNav: Cognitive Process Modeling for Object Goal Navigation with LLMs}

\author{
Yihan Cao$^{1,*}$ \\
Zheng Qin$^{3}$ 
\and
Jiazhao Zhang$^{2,*}$ \\
Qin Zou$^{4}$ 
\and
Zhinan Yu$^{1,*}$ \\
Bo Du$^{4}$
\and
Shuzhen Liu$^{1}$ \\
Kai Xu$^{1, \dagger}$
\and
\small{$^1$College of Computer Science and Technology, National University of Defense Technology}
\\
\small{$^2$ CFCS, School of Computer Science, Peking University }
\\
\small{$^3$Defense Innovation Institute, Academy of Military Sciences \quad
$^4$School of Computer Science, Wuhan University}
}
\usepackage{hyperref}
\usepackage[capitalize]{cleveref}
\begin{document}
\maketitle


\makeatletter
\def\blfootnote{\xdef\@thefnmark{}\@footnotetext}
\makeatother
\blfootnote{\noindent$^*$ Joint first authors; $^\dagger$ Corresponding author;\\Homepage: \url{https://yhancao.github.io/CogNav/}}
\begin{abstract}
Object goal navigation (ObjectNav) is a fundamental task in embodied AI, requiring an agent to locate a target object in previously unseen environments. This task is particularly challenging because it requires both perceptual and cognitive processes, including object recognition and decision-making. While substantial advancements in perception have been driven by the rapid development of visual foundation models, progress on the cognitive aspect remains constrained, primarily limited to either implicit learning through simulator rollouts or explicit reliance on predefined heuristic rules.
%
Inspired by neuroscientific findings demonstrating that humans maintain and dynamically update fine-grained cognitive states during object search tasks in novel environments, we propose \name, a framework designed to mimic this cognitive process using large language models. Specifically, we model the cognitive process using a finite state machine comprising fine-grained cognitive states, ranging from exploration to identification. Transitions between states are determined by a large language model based on a dynamically constructed heterogeneous cognitive map, which contains spatial and semantic information about the scene being explored.
%
%
Extensive evaluations on the HM3D, MP3D, and RoboTHOR benchmarks demonstrate that our cognitive process modeling significantly improves the success rate of ObjectNav at least by relative $14\%$ over the state-of-the-arts.




\end{abstract}
\vspace{-6pt}

%
%
\section{Introduction}
\label{sec:intro}

Navigating an unseen environment to find a target object, known as object goal navigation (ObjectNav)~\cite{habitatchallenge2023, li2022object}, is a fundamental task for embodied agents and has received increasing attention in the computer vision and robotics fields in recent years. To achieve ObjectNav, an agent must efficiently explore the environment, effectively reason about the spatial arrangement of target objects, strategically plan an optimal path to the goal, and accurately identify the target object upon approaching it. Consequently, ObjectNav is widely recognized as a challenging task that integrates both perceptual and cognitive processes~\cite{sun2024survey,chaplot2020object}.

\begin{figure}[t]
\begin{center}
  \includegraphics[width=1 \linewidth]{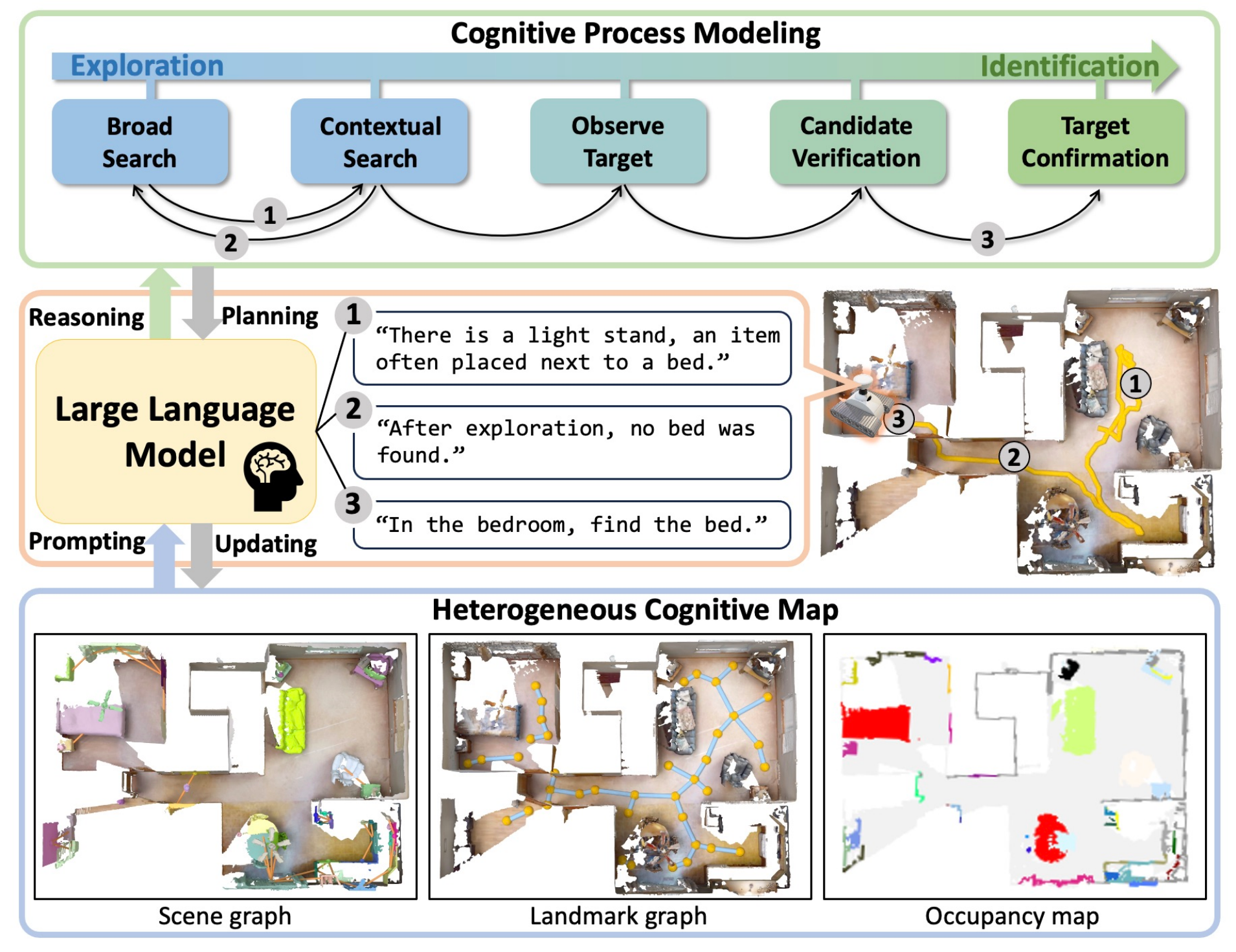}
\end{center}\vspace{-12pt}
   \caption{We propose to model the cognitive process of object goal navigation with an LLM. The LLM is leveraged to reason the transitions of a spectrum of states ranging from exploration to identification via being prompted by a heterogeneous cognitive map. The cognitive map is constructed online and dynamically updated/corrected through, again, prompting the LLM.}
\label{fig:teaser}
\vspace{-22pt}
\end{figure}

Recent advances in visual foundation models have significantly boosted the visual perception capabilities of embodied agents for scene understanding~\cite{radford2021learning, kirillov2023segment, liu2024grounding}, enabling rapid progress in zero-shot ObjectNav. In contrast, modeling the cognitive process of ObjectNav has so far received limited progress. One methodology~\cite{chaplot2020object, campari2020exploiting,dang2023search} involves implicit learning from demonstrations or rollouts generated by simulators~\cite{habitatchallenge2023,szot2021habitat,robothor}. However, due to the limited diversity of scenes and objects, these methods are known to suffer from poor generality in zero-shot settings.

Another approach~\cite{ramakrishnan2022poni,staroverov2022hierarchical,chen2024mapgpt} to modeling the cognitive process adopts pre-defined ObjectNav states and heuristic rules of state transition with prescribed thresholds~\cite{chaplot2020object, luo2022stubborn}. These explicit methods typically use two states, \textit{i.e.}, exploration and identification, demonstrating promising generality in handling diverse environments across various benchmarks. Moreover, explicitly decoupling the perceptual and cognitive processes leads to a hierarchical solution that naturally fits into zero-shot settings. However, the coarse two-state modeling fails to capture the full-fledged cognitive process of ObjectNav, such as utilizing scene context or making additional observations to verify the target object.


Neuroscientific evidence suggests that human brains maintain fine-grained cognitive states and continuously updates the states during object searches by, for instance, utilizing the context or making additional observations to verify~\cite{gabrieli1998cognitive,crivelli2023goal}. Inspired by this, two key questions are involved in modeling human cognitive process: \textit{How to set up a reasonable set of cognitive states and how to learn the state transitions to achieve efficient ObjectNav?} The latter requires extensive demonstrations to learn and is often difficult to attain cross-target or cross-scene generalization, especially when working on fine-grained states.

We propose CogNav, a framework that models the cognitive process of ObjectNav by leveraging the powerful reasoning abilities of large language models (LLMs). The LLM reasons the transitions of a carefully designed spectrum of cognitive states ranging from exploration to identification (Fig.~\ref{fig:teaser}). We show that an LLM is capable of effectively directing the ObjectNav process via \emph{deciding which state to take in the next}, by processing the abstract prompt of an online constructed cognitive map (see next). In particular, the LLM is prompted to examine the in-alignment of the cognitive map and reasons about the cognitive state transitions with realtime grounding of the target environment.

To maintain the incremental observations of the environment during navigation, we propose an online construction of a heterogeneous cognitive map that encompasses: 1) an LLM-friendly scene graph whose nodes represent detected objects with semantic descriptions and edges encode spatial relationships; 2) a navigation landmark graph, which records locations of interest in finding the target object; and 3) a top-view occupancy map for path planning. The heterogeneous cognitive map is incrementally constructed and can be dynamically updated or corrected through, again, prompting an LLM to ensure the map quality.

Our CogNav, with LLM-driven cognitive process modeling, leads to highly efficient and even human-like navigation behaviors without needing fine-tuning of the LLM. CogNav outperforms the previous state-of-the-arts (SOTA) with significant margins in both synthetic and real-world environments. In particular, it improves the SOTA ObjectNav success rate from $62\%$ to $72.5\%$ for the HM3D~\cite{ramakrishnan2021hm3d} benchmark, from $40.2\%$ to $46.6\%$ for MP3D~\cite{Matterport3D}, and from $47.5\%$ to $54.6\%$ for RoboTHOR~\cite{deitke2020robothor}.

To showcase the practical effectiveness of CogNav, \emph{we have implemented it with two real robot settings, a custom-built mobile robot and a Unitree quadruped}, demonstrating encouraging results in real-world tests; see the accompanying video submitted in the supplementary.

The promising results indicate that our method achieves arguably a full exploitation of the cognitive potential of LLMs in realizing spatial intelligence, saving the effort of training a policy model with numerous samples or trials and limited generality.
Our contributions include:
\begin{itemize}
    \item An effective cognitive process modeling for ObjectNav via exploiting the commonsense and spatial reasoning capability of LLMs.
    \item The design of fine-grained cognitive states and the prompts for grounded reasoning of state transitions with an LLM.
    \item A heterogeneous cognitive map representation that is constructed online and can be corrected by prompting an LLM to ensure high map accuracy.
\end{itemize}

\section{Related works}

\begin{figure*}[t]
\begin{center}
  \includegraphics[width=1 \linewidth]{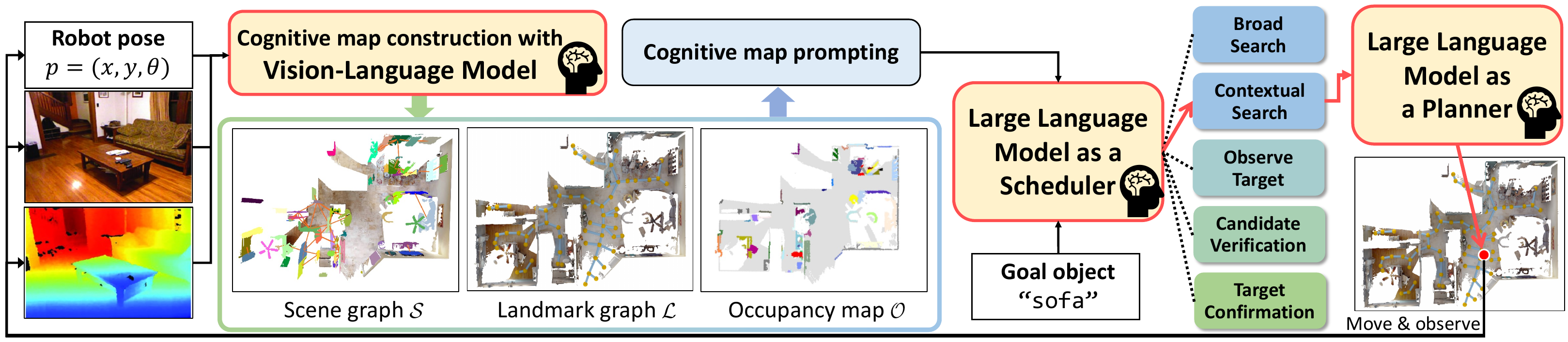}
\end{center}\vspace{-12pt}
   \caption{\textbf{Pipeline of CogNav}: Our method takes a posed RGB-D sequence as input and incrementally constructs an online cognitive map, comprising a scene graph, a landmark graph, and an occupancy map. We then perform cognitive map prompting by encoding cognitive information and goal object into a text prompt used to query the LLM to determine the next cognitive state. Based on the state, the LLM is queried again to select a landmark to guide the robot. A deterministic local planner is used to generate a path to the selected landmark.}
\label{fig:pipeline}
\vspace{-16pt}
\end{figure*}

\textbf{Object Goal Navigation.} 
Object goal navigation (ObjectNav) is a fundamental task for embodied agents. Existing ObjectNav approaches fall largely into two categories: end-to-end learning and modular methods \cite{chaplot2020object,ramakrishnan2022poni,zhou2023esc,yu2023l3mvn}. End-to-end methods learn to map observations to actions directly with reinforcement learning (RL) \cite{wijmans2019dd,ye2021auxiliary,dang2023search} or imitation learning (IL) \cite{ramrakhya2022habitat,ramrakhya2023pirlnav}. However, they face challenges in sim-to-real transfer and computational efficiency. Modular methods, as proposed in \cite{chaplot2020object}, decompose navigation into three key components: a mapping module for online environment representation \cite{zhang20233d,rudra2023contextual,staroverov2022hierarchical}, a policy module for long-term goal selection \cite{chaplot2020object,ramakrishnan2022poni,zhang20233d}, and a path planning module for action planning~\cite{sethian1996fast}. With the advancements in multi-modal representation learning \cite{radford2021learning}, ObjectNav now enables open-vocabulary object localization. Methods such as CoWs \cite{gadre2023cows} and ZSON \cite{majumdar2022zson} employ CLIP \cite{radford2021learning} to align visual inputs with target object categories although they often suffer from scenario-specific overfitting, limiting generality.
Unlike conventional object navigation, zero-shot object navigation focuses on locating objects of unseen categories while enhancing exploration efficiency. Recent studies \cite{yu2023l3mvn,zhou2023esc} integrate common sense knowledge from large language models (LLMs) to enable zero-shot decision making. Approaches such as OPENFMNAV \cite{kuang2024openfmnav} and TriHelper \cite{zhang2024trihelper} leverage LLMs to reason about scene information and select the next goal from a set of candidates. SG-NAV \cite{yin2024sg} utilizes LLMs to infer navigation points by analyzing structural relationships in scene graphs near frontier regions. Our proposed CogNav also employs an LLM-driven zero-shot framework but differs by explicitly modeling the cognitive process of ObjectNav. Unlike prior methods, it dynamically determines the next goal based on the evolving internal cognitive states of the agent driven by LLM reasoning.

\noindent\textbf{Large Language Models for ObjectNav.}
Large Language Models (LLMs) \cite{brown2020language,achiam2023gpt} have become pivotal in zero-shot navigation due to their strong reasoning and generalization capabilities. Recent studies \cite{yu2023l3mvn,kuang2024openfmnav,cai2024bridging,chen2024mapgpt,zhou2024navgpt} utilize LLMs to infer semantic relationships between objects and rooms, facilitating the optimal selection of navigation points from frontiers. Long et al.~\cite{long2024instructnav} propose a Dynamic Chain-of-Navigation framework to guide LLMs in handling diverse navigation instructions, while \cite{yin2024sg} employs a hierarchical Chain-of-Thought mechanism for scene graph reasoning, improving decision-making accuracy.

Since LLMs require text-based prompts for comprehension and analysis, constructing effective, LLM-friendly scene representations is crucial to extract relevant information from observations and convert them into LLM-readable descriptions. Most modular methods \cite{yu2023l3mvn,kuang2024openfmnav} adopt an explicit top-down semantic grid map \cite{chaplot2020object}, where frontiers are extracted to guide navigation. Zhang et al.~\cite{zhang20233d} enhance this approach by integrating 3D point-based scene representations as input. VoroNav \cite{wu2024voronav} introduces the Reduced Voronoi Graph to extract exploratory paths and planning nodes for efficient navigation. InstructNav \cite{long2024instructnav} leverages multi-sourced value maps to model key navigation elements. 
Meanwhile, the works of \cite{werby2024hierarchical,yin2024sg,loo2024open} utilize scene graphs as richer scene representations.
Building on the information richness of scene graphs and the improved reachable point representation of \cite{wu2024voronav}, we propose a heterogeneous map, a structured scene representation, to enhance LLM-driven scene understanding and navigation reasoning.

\section{Method}

\textbf{Task definition.}
The object goal navigation task requires the agent to navigate within an unknown environment to find an object in a given category $c$. At each time step $t$, the agent receives a posed RGB-D image $ I_t = \left \langle I_t^{\text{rgb}},I_t^{\text{depth}}, p_t \right \rangle$ where pose $p_t$ includes both location and orientation $p_t = (x_t, y_t, \theta_t)$.
The policy then predicts and executes an action $a_t \in \cA$, where $\cA$ consists of six actions: \texttt{move\_forward}, \texttt{turn\_left}, \texttt{turn\_right}, \texttt{look\_up}, \texttt{look\_down}, and \texttt{stop}. Given a limited time budget of 500 steps, the task succeeds if the agent stops within 1 meter of an object of $o$. By leveraging a foundation model \cite{radford2021learning,brown2020language,achiam2023gpt}, our method enables finding the target object in an open-vocabulary and zero-shot manner: the target category $c$ can be freely specified with text and the navigation system does not require any training or fine-tuning.
 
\noindent\textbf{Overview.}
Figure~\ref{fig:pipeline} provides an overview of CogNav.
Our method takes the target category $c$ and posed RGB-D frames as input and online construct a \emph{heterogeneous cognitive map} $\cC$, which consists of a semantic scene graph $\cS$, a landmark graph $\cL$ and a top-down occupancy map $\cM$ (Sec.~\ref{sec:cog_map}). We then employ a carefully designed prompting strategy to obtain landmark-centered descriptions of the cognitive map $\cC$ (Sec.~\ref{sec:cog_pro}), which includes both semantic and spatial information.
All prompting cognitive map information is passed to the large language model, to determine the current cognitive state, varying from broad exploration to target confirmation. With a given cognitive state, the large language model is prompted to adopt a specific skill to determine the next goal, such as exploration to another room or stopping at one landmark (Sec.~\ref{sec:cog_model}). This goal will be sent to the local planner for low-level action planning to execution.
\subsection{Cognitive Map Construction}
\label{sec:cog_map}

To comprehensively encode the scene environments, we online build a heterogeneous cognitive map $\cC$, which encompasses a scene graph $\cS$, a 2D occupancy map $\cM_t$, and a landmark graph $\cL_t$: 
\begin{equation}
    \cC_t = \{\cS_t, \cM_t, \cL_t\}.
\end{equation}
The scene graph provides detailed semantic information about instances, while the 2D occupancy map provides spatial information about the layout of the scenes. More importantly, the landmark graph serves as an anchor to aggregate both semantic and spatial information and offers navigable candidates for the robot to navigate.Details are elaborated as follows:

\noindent\textbf{Scene graph.} The scene graph $\cS_t=\left \langle \cN_t,\cE_t \right \rangle $ is composed of the instance nodes $\cN_t$ and spital informative edges $\cE_t$. Here each node $n_t \in \cN_t$ is an instance in the environment along with its detailed description of its semantics and spatial context and each edge $e_t \in \cE_t$ indicates a spatial relationship between two nodes in $\cN_t$ from a candidate set \{ 'next to', 'on top of', 'inside of', 'under', 'hang on'\}.  

During navigation, we employ an open-vocabulary segmentation approach~\cite{zhang2023simple} to obtain 3D instances along with semantic features by clustering instance segmentations in 3D space in an online manner. Specifically, at each time step $t$, objects are segmented from $I_t^{\text{rgb}}$ and their 2D instance masks are back-projected into a global 3D space using depth map $I_t^{\text{depth}}$ and pose $\theta_t$ to achieve 3D instance segmentation ${n_t^i}$. Subsequently, these segmented instances are merged with existing instances $\cN_{t-1}$ to update the instance nodes $\cN_{t}$ via DBSCAN clustering~\cite{khan2014dbscan}, following existing methods~\cite{gu2024conceptgraphs}. These instance nodes provide an initial representation of semantic and spatial information to construct the edges $\cE_t$.


However, directly obtaining segmentation results from the clustering algorithm is often noisy and prone to errors.
For a more accurate and detailed description of instance nodes and edges, we leverage the vision-language model (VLM, GPT-4v~\cite{achiam2023gpt} in our implementation) to infer the observation. We adopt a SoM~\cite{yang2023setofmark} technique by labeling 2D instances with projected $\cN_t$ and then reasoning the spatial and semantic information with outputting the relationship or room type.
For time efficiency, we perform the VLM inference every 10 steps, and only reason the newly fused instance and corresponding edges. We find this technique significantly removes the wrong instances and provides more accurate and detailed spatial information. \emph{We elaborate more on this in the supplementary.}

\noindent\textbf{Occupancy map.} To provide spatial scene information, such as explored areas, unexplored areas, and frontiers, we construct a top-down occupancy map $\cM_t \in \mathbb{R}^{M \times M \times 4}$ from $I_t$, following the approach in~\cite{chaplot2020object}. $M \times M$ defines the map size with a grid resolution of $5 \text{cm}$. The first three channels indicate occupied areas, explored regions, and the agent location, respectively. The fourth channel incorporates the projected 3D instance nodes $\cN_t$. The occupancy map $\cM_t$ is updated at each navigation step.

\noindent\textbf{Landmark graph.}
To anchor the scene graph and occupancy map information to navigation, we use a landmark graph $\cL_t = \{l_{t, i}\}$ to discrete the navigation area, 
where $l_{t,i}$ is a 2D location extracted from a Voronoi node graph~\cite{kalra2009incremental}.
The Voronoi graph partitions the navigation area based on proximity to a set of key locations. Specifically, we define two types of key locations:
\emph{frontier locations} representing the spatial information and \emph{instance locations} representing the object information. The frontier locations are clustered from the occupancy map following the method in \cite{yu2023l3mvn}.
The instance locations are the projection of instance nodes $\cN_t$ in the occupancy map.
Using these frontier and instance locations, we construct a Voronoi graph with a Generalized Voronoi Diagram~\cite{wu2024voronav}.
The landmark graph is updated every 10 frames, consistent with the scene graph update frequency. \emph{We elaborate more on this in the supplementary.}

\begin{figure}[t]
\begin{center}
  \includegraphics[width=0.95 \linewidth]{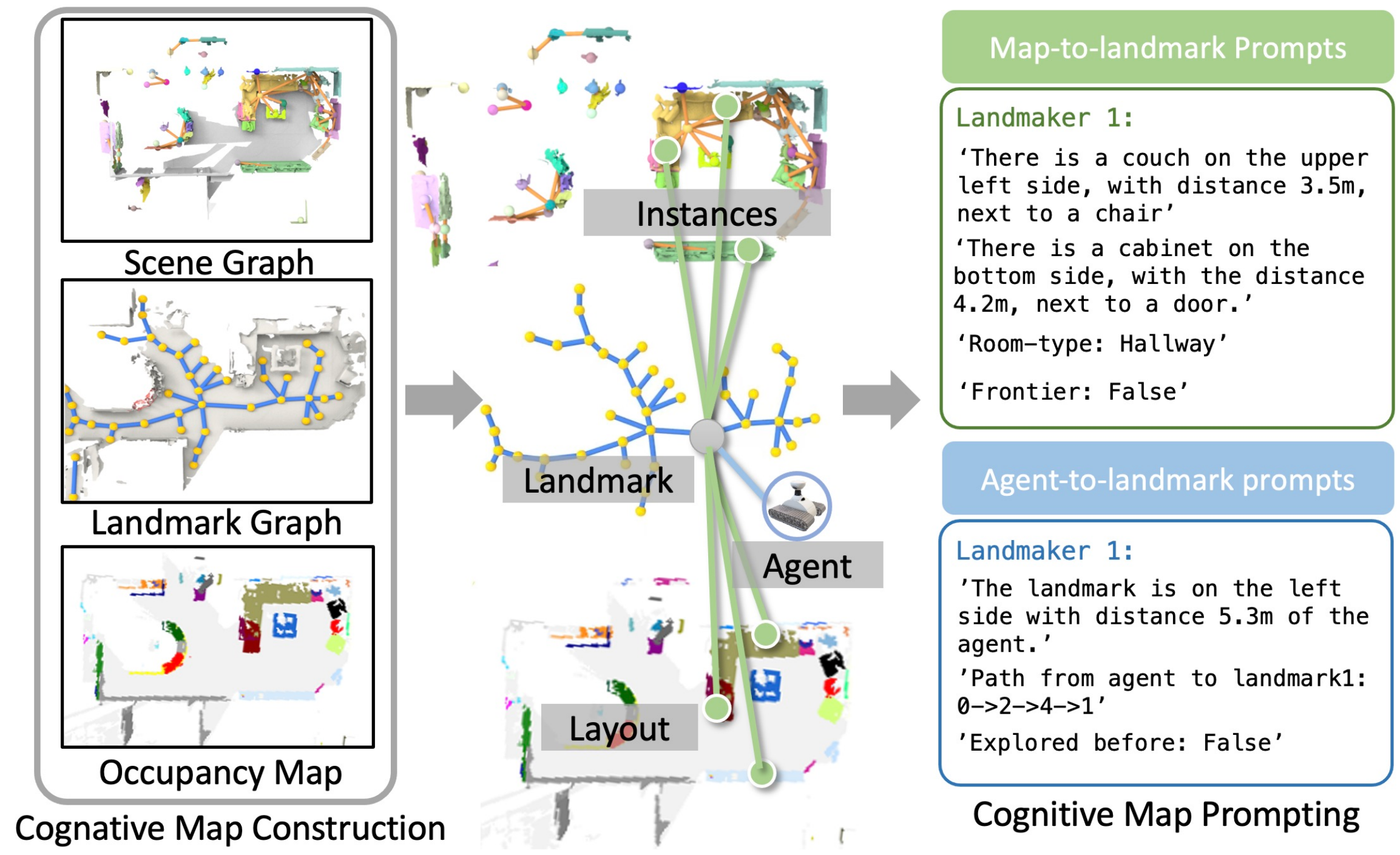}\vspace{-12pt}
\end{center}
   \caption{\textbf{Cognitive Map Prompting:} We encode knowledge from the cognitive map by constructing landmark-centered prompts encompassing both scene and agent information.}
\label{fig:cogmap}
\vspace{-16pt}
\end{figure}

\subsection{Cognitive Map Prompting}
\label{sec:cog_pro}




We encode the cognitive map $\cC_t$ knowledge in natural language by constructing landmark-centred prompts that embed scene information along with agent information $p_{1:t}$ into each landmark $\cL_t$:
\begin{equation}
    \begin{split}
    \cP_t = \{ & \text{PromptMap}(\cC_t, \cL_t), 
               \text{PromptAgent}(p_{1:t}, \cL_t) \}, \\
    \text{s.t.} \quad & p_t = (x_t, y_t, \theta_t).
    \end{split}
\end{equation}
Our cognitive map prompts (Figure~\ref{fig:cogmap}) comprise two types of information: (1) map-to-landmark prompts, which include details about neighboring objects and spatial relationships within the scene, e.g., next to a chair or near the scene boundary; and (2) agent-to-landmark descriptions, which convey navigational information relative to the landmarks, e.g., landmarks explored or close to the agent.



\noindent\textbf{Map-to-landmark prompts.} 
For each landmark $l_{t,i} \in \cL_t $, we integrate the scene information including surrounding objects along with their relationships, the room type and the frontier property.
We select object nodes $\cN_{t,l_i} \subseteq \cN_{t}$ within a certain distance threshold from a landmark $l_{t,i}$ in the scene graph $\cS_t$ as the surrounding objects. 
These nodes are prompted to incorporate their semantic information and spatial relationships to other nodes, thereby embedding the structural information of the scene graph to achieve a richer understanding of the surrounding environment. In addition, we also attach to the landmark room type information which is obtained from the VLM during scene graph construction (Sec.~\ref{sec:cog_map}). We then calculate the distance from the landmark to the nearest frontiers in the occupancy map and mark it as a frontier if the distance is less than 1 meter.

\noindent\textbf{Agent-to-landmark prompts.} 
%
We prompt the spatial relationship between the landmark and the agent. For each landmark $l_{t,i}$, we compute the landmark path from the agent to the landmark. Specifically, we calculate the distance between the landmark and the agent using the landmark graph $\cL_t$ via Dijkstra’s Algorithm. We also record whether this landmark has previously been reached by the agent, which helps to avoid redundant exploration.

\IncMargin{0.5em}
\begin{algorithm}[t]

\caption{Pipeline of CogNav}
\label{algo:fcog}
\SetKwInOut{AlgoInput}{Input}
\SetKwInOut{AlgoOutput}{Output}
\SetKwFunction{AgentObservation}{AgentObservation}
\SetKwFunction{CognitiveMapConstruction}{CognitiveMapConstruction}
\SetKwFunction{CognitiveMapPrompting}{CognitiveMapPrompting}
\SetKwFunction{CognitiveProcessModeling}{CognitiveProcessModeling}
\SetKwFunction{ObserveEnvironment}{ObserveEnvironment}
\SetKwFunction{LLM}{LLM}
\SetKwFunction{VLM}{VLM}
\SetCommentSty{\color{gray}\scriptsize} 
\SetKwComment{Comment}{}{}  

\definecolor{lightblue}{RGB}{69, 110, 196}
\definecolor{lightgreen}{RGB}{33, 140, 33}
\definecolor{nowblue}{RGB}{28, 27, 255}
\definecolor{lightgray}{RGB}{155, 155, 155}

\newcommand{\RightComment}[1]{{\hfill{\color{lightgray}$\triangleright$~#1}}}
\newcommand{\LeftComment}[1]{{{\color{lightgray}$\triangleright$~#1}}}


\AlgoInput{Target category $c$}
\AlgoOutput{ Task completion status}

\Repeat{$a_t$ is \text{stop}}{
    $\cI_{t} \gets$ \textit{AgentObservation}($t$)\; 
    $\cC_t\gets$ \textit{{\color{lightblue}VLM}CognitiveMapConst.($\cI_t$)}; \RightComment{Sec. 3.1} 
    
    $\cP_t \gets$ \textit{CognitiveMapPrompting}($\cC_t$, $c$); \RightComment{Sec. 3.2} 
    
    $state_t \gets$ \textit{{\color{lightblue}LLM}CogStateReasoning}($\cP_t$); \RightComment{Sec. 3.3} 
    
    \LeftComment{Given $state_t$, conduct different nav. strategies.}
    
    \lIf{$state_t = \text{BS}$}{\textit{goal} $\gets {\color{lightblue}LLM}LandmarkSelection(\cP_t,state_t, \cL_{t,\text{frontier}})$}
    \lIf{$state_t = \text{CS}$}{\textit{goal} $\gets {\color{lightblue}LLM}LandmarkSelection(\cP_t,state_t, \cL_{t,\text{relevent}})$}
    \lIf{$state_t = \text{OT}$}{\textit{goal} $\gets {\color{lightblue}LLM}LandmarkSelection(\cP_t,state_t, \cL_{t,\text{target}})$}
    \lIf{$state_t = \text{CV}$}{\textit{goal} $\gets {\color{lightblue}LLM}LandmarkSelection(\cP_t, state_t,\cL_{t,\text{verify}} )$}
    \lIf{$state_t = \text{TC}$}{\textit{goal} $\gets  \cL_{t,\text{target}} $}
    $a_t \gets Fast Marching Method(\textit{goal})$
}
\end{algorithm}
\begin{figure}[t]
\begin{center}
  \includegraphics[width=1 \linewidth]{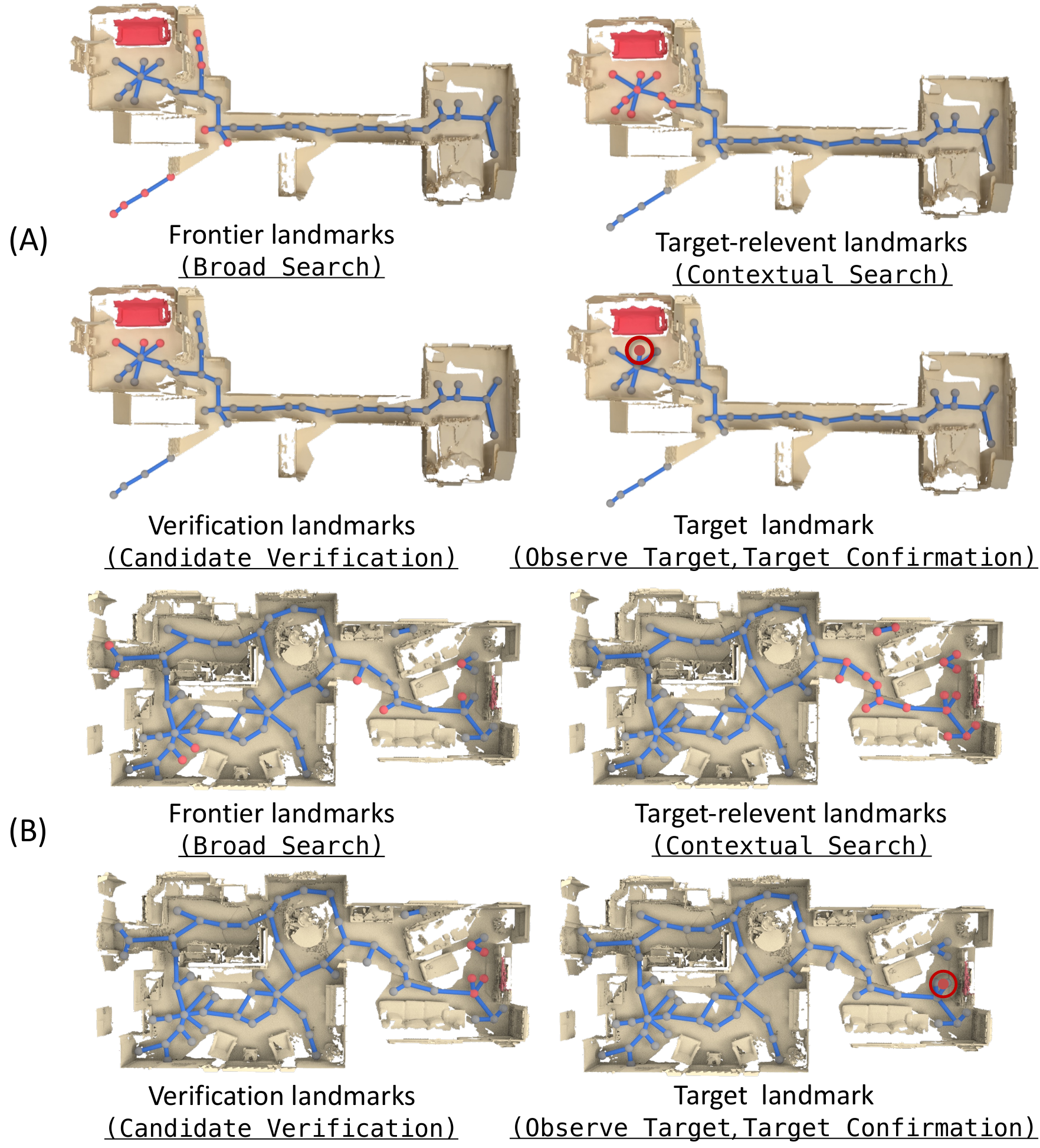}\vspace{-12pt}
\end{center}
   \caption{\textbf{State-related landmarks} (red dots) activated by different cognitive states. \emph{See the supplementary for a detailed explanation of landmarks.}}
\label{fig:landmarks}
\vspace{-20pt}
\end{figure}

\subsection{Cognitive Process Modeling} 
\label{sec:cog_model}

\begin{table*}[ht]
\centering
\scalebox{0.9}{
\begin{tabular}{lcccccccc}
\hline
\multirow{2}{*}{\textbf{Method}} &\multirow{2}{*}{\textbf{Open-Set}} &\multirow{2}{*}{\textbf{Zero-Shot}} & \multicolumn{2}{c}{\textbf{HM3D}} & \multicolumn{2}{c}{\textbf{MP3D}}  & \multicolumn{2}{c}{\textbf{RoboTHOR}} \\ \cmidrule(r){4-5} \cmidrule(r){6-7} \cmidrule(r){8-9} 
 & & & SR $\left ( \% \right )\uparrow$ & SPL $\left ( \% \right )\uparrow$  & SR $\left ( \% \right )\uparrow$ & SPL $\left ( \% \right )\uparrow$ & SR $\left ( \% \right )\uparrow$ & SPL $\left ( \% \right )\uparrow$   \\
\hline
SemEXP~\cite{chaplot2020object} &\ding{55}&\ding{55} &-- &-- &36.0 &14.4 &-- &--\\
PONI~\cite{ramakrishnan2022poni} &\ding{55}&\ding{55} &-- &-- &31.8 &12.1 &-- &-- \\ 
ZSON~\cite{majumdar2022zson}  &$\checkmark$ &\ding{55} &25.5 &12.6 &15.3 &4.8 &-- &-- \\ 
L3MVN~\cite{yu2023l3mvn} &\ding{55} &$\checkmark$ & 54.2 & 25.5 &34.9  &14.5 &41.2 &22.5 \\ 
ESC~\cite{zhou2023esc} &$\checkmark$ &$\checkmark$ & 39.2 & 22.3 &28.7  &11.2 & 38.1 & 22.2\\

VoroNav~\cite{wu2024voronav} &$\checkmark$ &$\checkmark$ & 42 & 26.0 &-- &-- &-- &-- \\

SG-NAV~\cite{yin2024sg} &$\checkmark$ &$\checkmark$ &54.0 & 24.9  & 40.2 &16.0 &47.5 &24.0 \\
OPENFMNAV~\cite{kuang2024openfmnav} &$\checkmark$ &$\checkmark$ &54.9 &22.4 &37.2 &15.7 & 44.1 &23.3\\
TriHelper~\cite{zhang2024trihelper} &$\checkmark$ &$\checkmark$ &62.0 &25.3 &-- &-- &-- &--\\
\hline
CogNav &$\checkmark$ &$\checkmark$ & \textbf{72.5} & \textbf{26.2}  & \textbf{46.6}  & \textbf{16.1} & \textbf{54.6} &\textbf{24.3}\\
\hline
\end{tabular}
}
\vspace{-2pt}
\caption{Comparison of success rate of different methods on HM3D~\cite{ramakrishnan2021hm3d}, MP3D~\cite{Matterport3D} and RoboTHOR~\cite{deitke2020robothor}.}
\vspace{-15pt}
\label{tab:benchmark}
\end{table*}

Inspired by the neuroscience evidence~\cite{crivelli2023goal} indicating that humans continuously update their cognitive states during object searches, we simulate this cognitive process by defining dense cognitive states, utilizing an LLM to determine the next cognitive state, and conducting the corresponding navigation strategy (See Algo.\ref{algo:fcog}).

\noindent\textbf{Cognitive states definition.}
To comprehensively encompass the cognitive states involved in object search, we define five cognitive states that progressively transition from exploration to identification:

\begin{itemize}
    \item \textbf{\textit{Broad Search}} (BS): Explore the environment without focusing on any specific object or room information related to the target object category $c$.
    \item \textbf{\textit{Contextual Search}} (CS): Investigate the rooms that are likely to contain the target or examine surrounding landmarks associated with objects closely related to the target.
    \item \textbf{\textit{Observe Target}} (OT): Recognize a potential target object based on the detection from the foundational model.
    \item \textbf{\textit{Candidate Verification}} (CV): Navigate to additional landmarks to verify the potential target object.
    \item \textbf{\textit{Target Confirmation}} (TC): Confirm the potential object as the target by analyzing the surrounding environment and navigation history, then approach it directly.
\end{itemize}


\noindent\textbf{Cognitive state transitions.} Despite these states providing a detailed modeling of the cognitive navigation process, it is challenging to determine which cognitive states are appropriate for the current situation. Unlike existing works~\cite{chaplot2020object,ramakrishnan2022poni,zhang20233d}, which often rely on thresholds to guide decision-making, we propose the use of LLMs as a powerful tool to analyze prompts from the cognitive map and determine the most suitable cognitive states for the current situation. We find that this design improves performance (Table~\ref{tab:state}) and demonstrates adaptability across different object categories (Figure~\ref{fig:transition}).

\noindent\textbf{State-guided navigation strategy.}
Given the cognitive states, we design corresponding navigation strategies for the agent to execute. These strategies are developed based on either the Large Language Model (LLM) to analyze the current landmark-centered scene information and select the most suitable landmarks to guide the agent's navigation.

Firstly, the agent is required to perform a broad search (\textit{Broad Search}) of the scene environment to maximize its understanding of the scene. In this process, we utilize LLM to select a frontier landmark $\cL_{t,\text{frontier}}$, which effectively guides the agent in exploring the environment to gather scene information. Once the agent observes areas associated with the target object, such as a room that may contain the target object (\textit{Contextual Search}), the agent is required to meticulously examine the relevant landmarks $\cL_{t,\text{relevant}}$.

During exploration, when the cognitive map identifies an object that could potentially be the target object (\textit{Observe Target}), the agent is required to approach the target object. Throughout the approach, the cognitive map is continuously updated. However, the target object prediction may remain inconsistent, thus prompting the LLM to transition the state to \textit{Target Verification}. Subsequently, the LLM predicts landmarks $\cL_{t,\text{verify}}$ near the potential target object to gather further observations.

Finally, when the LLM believes the cognitive map strongly supports the observed object is the target (\textit{Target Confirmation}), it will directly guide the agent to the target landmarks $\cL_{t, \text{target}}$. Note that, the cognitive map is online updated, which makes CogNav supports recovery from wrong predictions, such as from Target Verification to Broad Search. During navigation, we employ the Fast Marching Method (FMM)~\cite{sethian1996fast} to guide agent to the landmark goal. Once the agent arrives at the landmark goal and the current cognitive state is Target Confirmation, the agent outputs \texttt{STOP} to end the search. 
\emph{We elaborate more on this in the supplementary.}
\section{Experiment}
We evaluate our method by comparing it with several baselines in a simulated environment~\cite{szot2021habitat}. We also conduct ablation studies to validate our design choices. We also test our method in real-world robot platforms. 
\subsection{Experiment Setup}
\noindent\textbf{Datasets.} Our simulated experiments are conducted on three datasets:
1) The HM3D~\cite{ramakrishnan2021hm3d} dataset, a large-scale 3D indoor scene dataset for simulating embodied AI tasks. It includes 20 high-fidelity reconstructions of entire buildings and 2K validation episodes for object navigation tasks across six goal object categories;
2) The MP3D~\cite{Matterport3D} dataset, a comprehensive 3D indoor environment dataset created from real-world building scans. It contains 11 indoor scenes and 1.8K validation episodes for object navigation tasks across 20 goal object categories;
3) The RoboTHOR~\cite{deitke2020robothor} dataset, contains 1.8K validation episodes on 15 validation environments with 12 goal object categories.

\begin{figure*}[t]
\begin{center}
  \includegraphics[width=1 \linewidth]{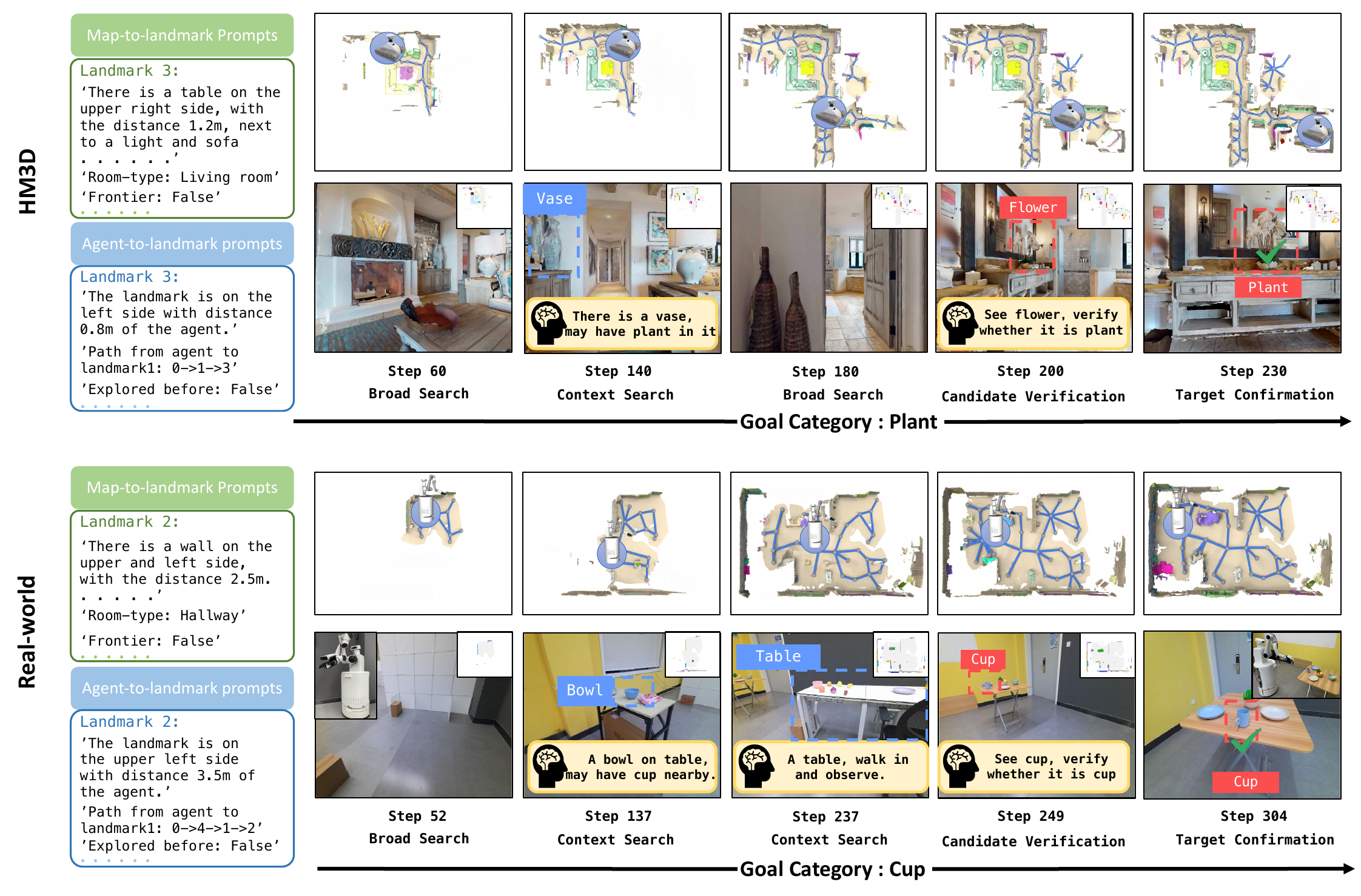}
\end{center}
\vspace{-20pt}
   \caption{\textbf{Navigation process visualization of CogNav.} We provide visual results of navigation process for one synthetic scene and one real-world one. Cognitive maps encode scene information and facilitate landmark prompting, enabling the agent to explore environments efficiently and identify target objects accurately. \emph{More results can be found in the supplementary.}}
\label{fig:gallery}
\vspace{-10pt}
\end{figure*}

\noindent\textbf{Evaluation Metrics.}
We use three metrics~\cite{anderson2018evaluation} to evaluate object navigation performance:
1) Success Rate (SR): The percentage of navigation episodes in which the robot successfully reaches within a certain distance of the target object;
2) Success weighted by Path Length (SPL): The success rate normalized by the optimal path length, favoring agents that navigate efficiently. Shorter paths are preferred over longer, suboptimal routes;
3) Distance to Goal (DTG): The final distance between the agent and the target object at the end of the navigation episode.

\noindent\textbf{Implementation Details.} For each episode, we set the maximum navigation step to 500. The observation of the agent is $640 \times  480$ RGB-D images,, with depth values ranging from $0.5$m to $5$m. The camera of agent is $0.9$m above the ground, with each forward action advancing $0.25$m and each rotation covering $30^{\circ}$. The occupancy map is configured as $960 \times  960$ with a resolution of $0.05$m. For instance detection and segmentation, we utilize OpenSEED~\cite{zhang2023simple} as our 2D foundation model. In the cognitive process, GPT-3~\cite{brown2020language} is used as the large language model (LLM), and GPT-4v~\cite{achiam2023gpt} serves as the vision-language model (VLM) for decision-making during navigation.

\noindent\textbf{Baselines.} We compare CogNav with several recent works, focusing on open-set and zero-shot object navigation baselines to verify the effectiveness of our framework.
Baselines are categorized into 'Open-Set' and 'Zero-Shot' methods:
'Open-Set' indicates that the approach can detect and navigate to any object category as specified and 'Zero-shot' indicates the approach can find and navigate toward target objects it has never encountered during training. For closed-set and unsupervised approaches, we cover SemExp~\cite{chaplot2020object} and PONI~\cite{ramakrishnan2022poni}. For open-set and unsupervised approach, we choose ZSON~\cite{majumdar2022zson}. For closed-set and zero-shot method, we choose L3MVN~\cite{yu2023l3mvn}. For open-set and zero-shot methods, we cover ESC~\cite{zhou2023esc}, VoroNav~\cite{wu2024voronav}, SG-NAV~\cite{yin2024sg}, OPENFMNAV~\cite{kuang2024openfmnav} and TriHelper~\cite{zhang2024trihelper}. 

\subsection{Quantitive Experiments}


We evaluate our approach on the HM3D (val), MP3D (val) and RoboTHOR (val) datasets with other baselines, including both open-set and zero-shot methods. The results are presented in Table~\ref{tab:benchmark}. Our approach achieves state-of-the-art performance on both datasets, outperforming other methods by a significant margin (a category-wise comparison can be found in  Table~\ref{tab:category}). For the success rate (SR) metric, our approach achieves a $10.5\%$ improvement on HM3D, a $6.4\%$ improvement on MP3D and a $7.1\%$ improvement on RoboTHOR.
Note that, the improvement in SR is more pronounced than in SPL. This is because our cognitive process modeling enables the agent to recover from incorrect predictions (\textit{e.g.}, candidate verification), significantly enhancing the success rate. However, during candidate verification, the agent must approach the goal object and observe it from multiple views. Consequently, this may increase the navigation path length. Similar findings can be found in Table.~\ref{tab:state}.
\emph{The efficiency and time analysis of our method are given in the supplementary.}


\begin{table}[t]
    \centering
    \scalebox{0.9}{
    \setlength{\tabcolsep}{0.6mm}{
    \begin{tabular}{lccccccc}
        \hline
        \textbf{Method} & \textbf{sofa} & \textbf{bed} & \textbf{chair} & \textbf{plant} & \textbf{toilet} & \textbf{tv\_monitor} & \textbf{average} \\
        \hline
        L3MVN~\cite{yu2023l3mvn}   & 50.1   & 52.9  & 51.6 & 46.4 &41.5 &54.2 & 49.5 \\
        TriHelper~\cite{zhang2024trihelper}   & 58.9  & 57.1 & 58.6 & 58.3 & 52.3 & 57.4 & 57.1  \\
        \hline
        CogNav  & \textbf{67.0} & \textbf{67.9} & \textbf{73.4} & \textbf{73.1} & \textbf{72.6} & \textbf{74.0} & \textbf{71.3}   \\
        \hline
    \end{tabular}
    }
    }
    \caption{The success rate of each category on HM3D~\cite{ramakrishnan2021hm3d}. The HM3D dataset contains six object categories in total. The 'average' represents the mean success rate across these six categories.}
    \label{tab:category}
\vspace{-18pt}
\end{table}
\begin{figure}[t]
\centering
\begin{overpic}
[width=\linewidth]
{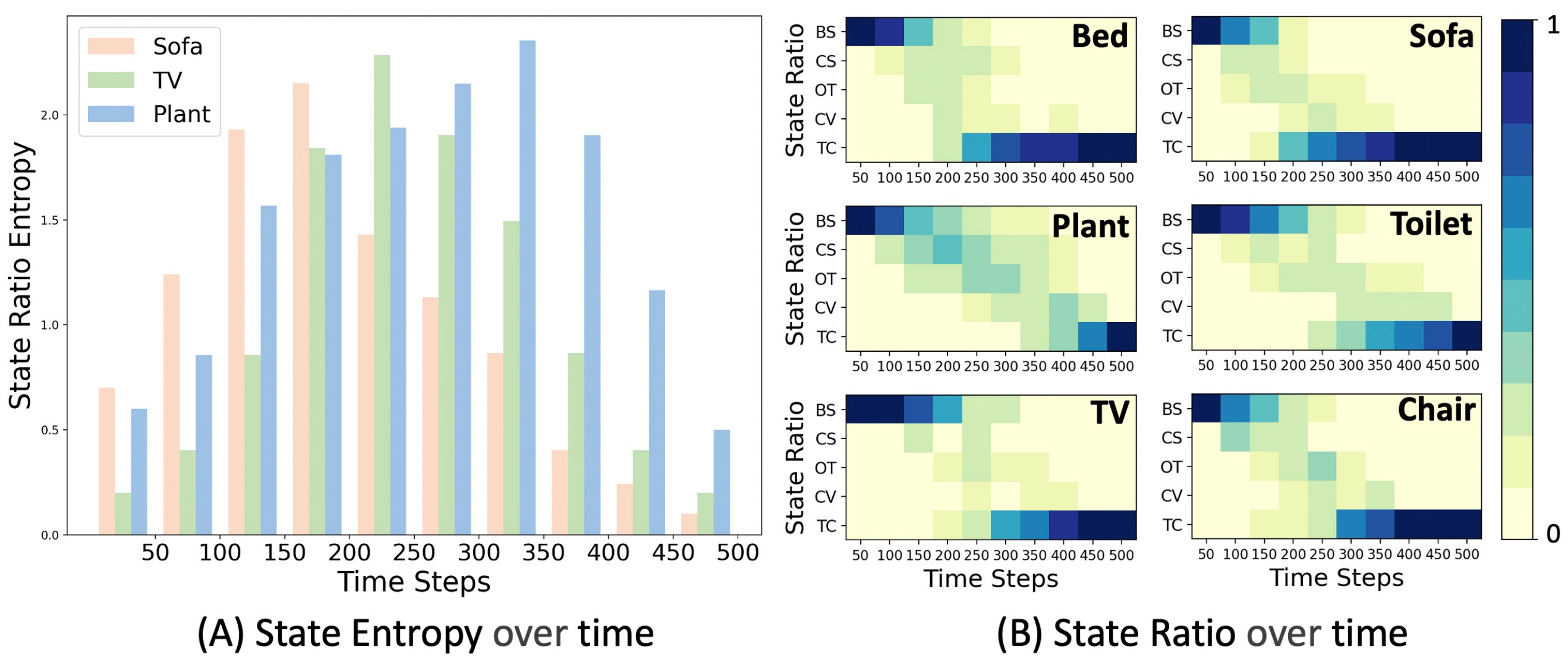}
\end{overpic}
\vspace{-6mm}
\caption{ State ratio and entropy over time steps. State entropy represents cognitive state discrepancy. The higher the entropy, the greater the variability in the states of locating the target object. The state ratio represents the proportion of selecting specific states.
}
\label{fig:transition}
\vspace{-18pt}
\end{figure}



We further conduct a breakdown experiment to investigate the cognitive state modeling during navigation. Specifically, we plot the state ratio entropy and state ratio across different time horizons (Figure~\ref{fig:transition}). Notably, when searching for a \texttt{sofa} or a \texttt{bed}, the states typically follow an orderly sequence from exploration to identification. In contrast, when searching for a \texttt{plant} or a \texttt{toilet}, the states are distributed irregularly, with frequent transitions between states, resulting in a longer search process. 
This comparison demonstrates that our cognitive process model adapts to the complexity of searching different object categories. 

\subsection{Qualitative Experiments}
We visualize the navigation process of CogNav in Figure~\ref{fig:gallery}. The upper example (rows 1-2) illustrates the task of finding a plant on HM3D, demonstrating how the agent transitions between states by utilizing the large language model’s (LLM) analysis of the scene and executing appropriate strategies to find the target object.
We also deploy our method on two real robots for object navigation in a real-world environment. \emph{A detailed description of the robot settings is in the supplementary.}
The lower example (rows 3-4) shows the task of finding a cup in a real-world scene, demonstrating that CogNav can perform open-set object navigation tasks in real environments.

\subsection{Ablation Study}
In Table~\ref{tab:state}, we compare the effects of different states in cognitive process modeling. Exploration in an unknown environment requires BS states, so we remove other states to verify the role of the remaining four states. As shown in the table, each state is critical to successfully finding the target object. Note that, removing the candidate verification state (row 4) decreases $3.7\%$ in SR but increases $7.8\%$ in SPL, due to the increased navigation path length when the agent needs to visit different viewpoints of the target object for observation and confirmation. We referenced methods that consider some of the five states and found that the more states considered, the better the navigation performance. Compared to methods that consider the same states, our superior results also demonstrate that the form of a cognitive map for prompting is more effective in enabling LLMs to understand and reason.

We also perform an ablation study to examine the effectiveness of different cognitive prompt components in our approach. Results are shown in Table~\ref{tab:cogmap}. Adding edges with room-type in map-to-landmarks prompts (rows 1-3) significantly improves performance by incorporating relationships with surrounding objects and the room in which the agent is located. Additionally, rows 4-5 confirm the effectiveness of prompting spatial information and navigation history for LLM inference, respectively.

\begin{table}[t]
\centering
\scalebox{0.85}{
\setlength{\tabcolsep}{0.3mm}{
\begin{tabular}{ccccccccc}
\hline
\multirow{2}{*}{\textbf{Method}} &\multirow{2}{*}{\textbf{BS}} & \multirow{2}{*}{\textbf{CS}} & \multirow{2}{*}{\textbf{OT}} & \multirow{2}{*}{\textbf{CV}} & \multirow{2}{*}{\textbf{TC}} & \multicolumn{3}{c}{\textbf{HM3D}}  \\ \cmidrule(r){7-9}
& & & & & &SR $\left ( \% \right ) \uparrow$ &SPL $\left ( \% \right ) \uparrow$  &DTG $\left ( m \right ) \downarrow$ \\
\hline
L3MVN~\cite{yu2023l3mvn} &$\checkmark$ &\ding{55} &$\checkmark$ &\ding{55} &\ding{55} &54.2 &25.5 &4.427\\
\cellcolor{gray!20} CogNav& \cellcolor{gray!20} $\checkmark$ &\cellcolor{gray!20} \ding{55} &\cellcolor{gray!20} $\checkmark$ &\cellcolor{gray!20} \ding{55} &\cellcolor{gray!20} \ding{55} &\cellcolor{gray!20} 55.2 &\cellcolor{gray!20} 24.6 &\cellcolor{gray!20} 3.320\\
\cellcolor{gray!20} CogNav&\cellcolor{gray!20} $\checkmark$ &\cellcolor{gray!20} $\checkmark$ &\cellcolor{gray!20} $\checkmark$ &\cellcolor{gray!20} \ding{55} &\cellcolor{gray!20} \ding{55} &\cellcolor{gray!20} 57.6&\cellcolor{gray!20} 24.9&\cellcolor{gray!20} 2.708 \\
SG-Nav~\cite{yin2024sg}&$\checkmark$ &\ding{55} &$\checkmark$ &$\checkmark$ &\ding{55}&54.0&24.9&--\\ 
\cellcolor{gray!20} CogNav&\cellcolor{gray!20} $\checkmark$ &\cellcolor{gray!20} $\checkmark$ &\cellcolor{gray!20} $\checkmark$ &\cellcolor{gray!20} $\checkmark$ &\cellcolor{gray!20} \ding{55}&\cellcolor{gray!20} 67.2&\cellcolor{gray!20} 25.3&\cellcolor{gray!20} 1.983\\ 
TriHelper~\cite{zhang2024trihelper} &$\checkmark$ &\ding{55} &$\checkmark$ &\ding{55} &$\checkmark$ &62.0&25.3&3.873\\ 
\cellcolor{gray!20} CogNav&\cellcolor{gray!20} $\checkmark$ &\cellcolor{gray!20} $\checkmark$ &\cellcolor{gray!20} $\checkmark$ &\cellcolor{gray!20} \ding{55} &\cellcolor{gray!20} $\checkmark$ &\cellcolor{gray!20} 68.8&\cellcolor{gray!20} \textbf{33.6}&\cellcolor{gray!20} 1.423\\ 
\hline
\cellcolor{gray!20} CogNav&\cellcolor{gray!20} $\checkmark$ &\cellcolor{gray!20} $\checkmark$ &\cellcolor{gray!20} $\checkmark$ &\cellcolor{gray!20} $\checkmark$ &\cellcolor{gray!20} $\checkmark$ &\cellcolor{gray!20}  \textbf{72.5} &\cellcolor{gray!20}  26.2 & \cellcolor{gray!20} \textbf{1.255}\\
\hline
\end{tabular}
}
}
\caption{Ablation study of leveraging different cognitive states. BS, CS, CV and TC represent Broad Search, Contextual Search, Candidate Verification and Target Confirmation in Cognitive Process Modeling, respectively on HM3D~\cite{ramakrishnan2021hm3d}.}
\vspace{-5pt}
\label{tab:state}
\end{table}
\begin{table}[t]
\centering
\scalebox{0.9}{
\setlength{\tabcolsep}{1mm}{
\begin{tabular}{lccc}
\hline
\multirow{2}{*}{\textbf{Method}} & \multicolumn{3}{c}{\textbf{HM3D}}  \\ \cmidrule(r){2-4}
&SR $\left ( \% \right ) \uparrow$ &SPL $\left ( \% \right ) \uparrow$  &DTG $\left ( m \right ) \downarrow$ \\
\hline
Only object categories &63.3 &21.3 &2.651 \\
Remove edges &68.2 &21.8 &1.895\\
Remove room-type &71.6 &25.2 &1.466\\
\hline
Remove spatial information &64.5 &21.5 &1.752  \\
Remove navigation history &67.8 &22.3 &1.523\\
\hline
Complete Prompt & \textbf{72.5} & \textbf{26.2} & \textbf{1.255} \\ 
\hline
\end{tabular}
}
}
\caption{Ablation study of using different prompt strategies on HM3D~\cite{ramakrishnan2021hm3d} by removing respective parts from the map-to-landmark (rows 1-3) and agent-to-landmark prompts (rows 4-5).}
\vspace{-18pt}
\label{tab:cogmap}
\end{table}

\section{Conclusion and Future Work}

In this paper, we have proposed \name, a zero-shot framework that models the cognitive process for object goal navigation using large language models (LLMs). Our method constructs a heterogeneous cognitive map and models the cognitive process with a finite state machine composed of cognitive states range from exploration to identification to navigate the agent. Our method achieves state-of-the-art (SOTA) performance among all methods and validates the feasibility with a robot to navigate an object in a real-world scene. 

\noindent\textbf{Limitations and Future Work.}
Despite the great performance, \name still has several limitations: (1) \name relies on the detection and segmentation results from a 2D foundation model. If the model consistently fails to detect the target object, \name cannot locate the target within the time budget.
(2) Currently, \name is limited to object-goal navigation and does not support navigation based on free-form language descriptions or specific navigation instructions. However, given the powerful generalization abilities of LLMs and vision-language models (VLMs), our approach can be further extended to handle navigation tasks with unlimited input formats in future work.

\section{Acknowledgements}
This work was supported in part by the NSFC (62325211, 62132021, 62402516), the Major Program of Xiangjiang Laboratory (23XJ01009), Key R\&D Program of Wuhan (2024060702030143).
{
    \small
    \bibliographystyle{ieeenat_fullname}
    \bibliography{main}
}

\end{document}


\maketitle

This document offers a detailed explanation of our approach, along with additional experimental results, presented in the following structure.
\begin{itemize}
    \item Technical details (\S\ref{supsec:pipeline})
    \item More Experiments (\S\ref{supsec:experiment})
\end{itemize}
\section{Technical details}
\label{supsec:pipeline}
\subsection{Notation in paper}
We provide a lookup table of notations and their description mentioned in this paper in Table~\ref{tab:notation} for reference.
\subsection{Cognitive Map in paper Sec.3.1.}
\begin{figure}[t]
\centering
\begin{overpic}
[width=\linewidth]
{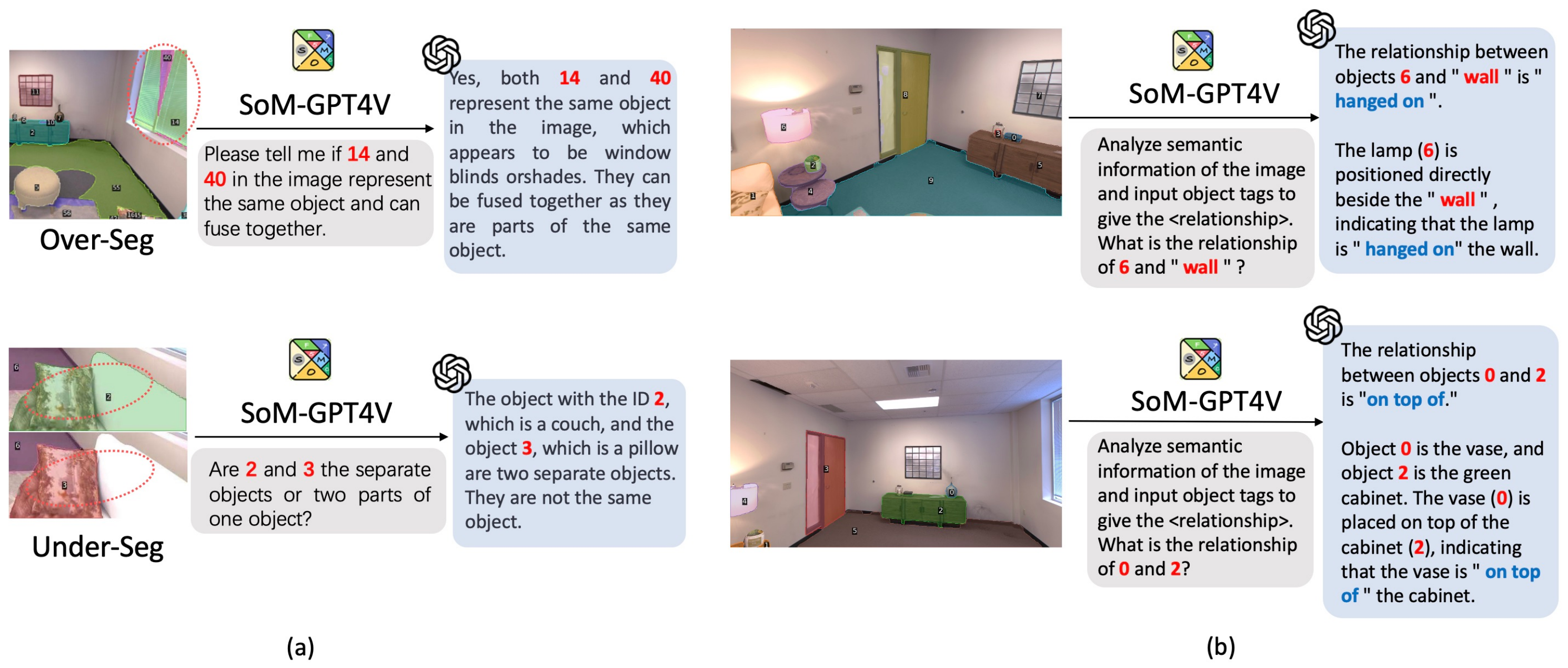}
\end{overpic}
\caption{\textbf{Fusion and Relationship Query in Cognitive Map.} We use GPT-4v prompted by SoM image to solve the problem of fusion and relational edges in Cognitive Map. Column (a) shows how to confirm whether two partitions are the same instance. Column (b) shows the relationship generation between two adjacent instances. 
}
\label{fig:cogvlm2}
\vspace{-8pt}
\end{figure}

\noindent\textbf{Details of Scene Graph.}
The scene graph $\cS_t$ comprises instance nodes $\cN_t$ and relationship edges $\cE_t$.
The construction of node $\cN_t$ is based on an online open-vocabulary segmentation method \cite{gu2024conceptgraphs} to obtain the 3D instances. Unlike \cite{gu2024conceptgraphs}, we employ a more accurate and stable detection and segmentation framework, OpenSEED \cite{zhang2023simple} as our foundation model and replace point cloud object representation with a voxel grid for clearer object boundaries and better alignment with the 2D occupancy map $\cM$. At each time $t$, we detect and segment objects from $I_t^{rgb}$ and reconstruct a voxel grid in global coordinates by $I_t^{depth}$ and $\theta_t$ synchronously. 
Each node $\cN_t$ contains the 3D voxel grid coordinates, the detected category, the semantic features of the instance, and the semantic features corresponding to the detected category.
The fusion strategy in \cite{gu2024conceptgraphs}, which simply matches and fuses newly detected objects at time $t$ with the object nodes $\cN_{t-1}$ from previous frames, may suffer from under-segmentation. This limitation can lead to under-detection of small objects, posing challenges for navigation tasks.
To address over-segmentation and under-segmentation issues, we utilize Vision-Language Models (VLMs). To uniquely represent each instance $n \in \cN_t$ in the scene graph, we utilize SoM \cite{yang2023setofmark}, which masks and marks the instance with a node number at the center of its mapping in the image. For ambiguous cases where it is unclear whether two instances should be merged, we use SoM to generate an image containing both instances, as illustrated in Figure~\ref{fig:cogvlm2}(a).
Objects that do not match any existing instances from previous frames are treated as new instances and are added as separate nodes in $\cN_t$.

After constructing $\cN_t$, we leverage VLMs with spatial context of objects to update edges $\cE_t$. We only update edges between object fused or created at time $t$ with surrounding objects considering the time cost. Edges that can be reasoned from spatial context will be directly computed like 'on top of' or 'under', edges that cannot be reasoned from spatial context are obtained by querying the VLMs. We use SoM \cite{yang2023setofmark} to mask and mark the two instances whose relationship needs to be queried on an image. The VLM is then prompted with this image to determine the relationship between the two objects, considering observations from two different viewpoints. The VLM selects one of the candidate set \{ 'next to', 'on top of', 'inside of', 'under', 'hang on'\}, with an additional 'none' indicating that two nodes have no relationship. We show the edge generation by VLM in Figure~\ref{fig:cogvlm2}(b).
\begin{table*}[h!]
\centering

\begin{tabular}{lll}
\hline
\textbf{Notation} & \textbf{Type/Unit} & \textbf{Description} \\
\hline
$t$ & number & a time step\\
$c$ & string & the category of the target object\\
$\mathcal{I}^{rgb}_t$ & Matrix (640 $\times$ 480 $\times$ 3) & a RGB image by agent observation at $t$ \\
$\cI^{depth}_t$ & Matrix (640 $\times$ 480) & depth image by agent observation at $t$\\
$p_t$ & $ \left \langle x_t, y_t, \theta_t \right \rangle  $ & current pose of agent, including the planar coordinates and rotation angle at $t$\\
$\cI_t$  & $ \left \langle I_t^{\text{rgb}},I_t^{\text{depth}}, p_t \right \rangle$ & posed RGB-D image at $t$\\
$a_t$ & an action & an action that agent executes in scene\\
$n_t$ & dict  & an instance dict in scene, including its semantic and spatial context at $t$ \\
$e_t$ & a spatial relationship between two $n_t$ & a relationship in [ 'next to', 'on top of', 'inside of', 'under', 'hang on' ] \\
$\cN_t$ & $\{n_t\}$ & the set of instance $n_t$ in scene at $t$\\
$\cE_t$ & $\{e_t\}$ & the set of relationship $e_t$ between two nodes in $\cN_t$ at $t$\\
$\cS_t$ & $\left \langle \cN_t,\cE_t \right \rangle$& the scene graph composed of $\cN_t$ and $\cE_t$ at $t$\\
$M$ & number & The map size with a grid resolution of 5cm\\
$\cM_t$ & Matrix(M $\times$ M $\times$ 4) & the top-down occupancy map at $t$ \\
$l_{t, i}$ & $ \left \langle x_{t,i}, y_{t,i}\right \rangle $ & a 2D location coordinate extracted from a Voronoi node graph at $t$\\
$\cL_t$ & $\{l_{t, i}\}$ & a landmark graph extracted from the Voronoi node graph at $t$ \\
$\cC_t$ &  $ \left \langle \cS_t,\cM_t,\cL_t \right \rangle $ & a coordinate-aligned heterogeneous cognitive map at $t$  \\
$\cD_t$ & dict & a dict mapping the leaf node to $n_t$ or a frontier area \\
$\gamma_{t,i}$ & string & the room type $l_i$ belongs to at $t$ \\
$\delta_{t,i}$ & bool & the frontier property of $l_i$ at $t$ \\
$l_{t,now}$ & $ \left \langle x_{t,now}, y_{t,now}\right \rangle $ & a 2D coordinate which the agent locates at \\
$n_c$ & dict & a node in $\cS_t$ whose object category is c \\
$\cL_{t,frontier}$ & set & The subset of $\cL_t$ whose $\delta_{t,i}$ is True \\
$\cL_{t,relative}$ & set & The subset of $\cL_t$ who are relative to the target object \\
$\cL_{t,target}$ & set & The subset of $\cL_t$ who are directed to $n_c$ \\
$\cL_{t,verify}$ & set & The subset of $\cL_t$ who are directed to $n_c$ and not explored before \\
\hline
\end{tabular}
\caption{Summary of Mathematical Notation}
\label{tab:notation}
\end{table*}

\noindent\textbf{Details of Landmark Graph.}
At time $t$, we construct a landmark graph $\cL_t$ using a semantic occupancy map $\cM_t$ to guide agent navigation. The landmark graph $\cL_t$ is derived from a processed Reduced Voronoi Diagram(RVD) extracted from the occupancy map $\cM_t$.
We first construct a top-down semantic occupancy map $\cM_t$ using $I_t$, similar to \cite{chaplot2020object}.
It is represented as a $ M\times M \times3 $ tensor with $ M\times M $ as map size and 3 channels including an obstacle and an explored channels from $I_t^{depth}$, with an semantic map, where each mask is aligned with the voxel grid of nodes $\cN_t$ and each channel stores the unique label of the aligned node. The semantic occupancy map is updated in real-time with each frame during navigation.

After constructing the semantic occupancy map, we refer \cite{yu2023l3mvn} to cluster frontiers and refer \cite{wu2024voronav} to extract a Generalized Voronoi Diagram(GVD) by skeletonizing the traversible areas. We remove redundant nodes and retain only the intersections and all leaf nodes in GVD to generate a Reduced Voronoi Diagram(RVD), different from the reduction strategy in \cite{wu2024voronav}, which only keeps the frontier end points. We propose a landmark-instance mapping $\cD_t$ such that each leaf node in RVD corresponds to the node label or frontier area in semantic map pointed to by the edge connecting this leaf node in the graph.
We fuse leaf nodes that point to the same object or frontier and are close to each other on RVD, and ensure that at least two leaf nodes point to the same object, which helps in the candidate verification process in Sec.3.3. The landmark graph $\cL_t$ at time $t$ is generated by fusing leaf nodes and subsequent reduction strategy on RVD, where each node $ l \in \cL_t$ is a reachable landmark that agent can navigate to as a long-term goal. Our GVD has a residual situation due to the fact that it only collects information about the scene from a single view, and therefore needs to update with the navigation process, unlike \cite{wu2024voronav} that uses panorama views to be able to build a complete GVD in the area centered on the agent.

\subsection{Cognitive Map Prompting in paper Sec.3.2.}
\noindent\textbf{Map-to-landmark prompts.}
For each landmark $l_{t,i} \in \cL_t $, we integrate the scene information to understand the environment of landmark $l_{t,i}$. The information includes surrounding objects along with their relationships, the room type $\gamma_{t,i}$ and the frontier property $\delta_{t,i}$.
We select object nodes $\cN_{t,l_i} \subseteq \cN_{t}$ within a certain distance threshold from landmark $l_{t,i}$ in the scene graph $\cS_t$ as the surrounding objects. Each node is prompted not only with its own information, including category name and semantic information but also with the information of its adjacent nodes within $\cS_t$, linked by edges that define the relative relationship. This prompting approach leverages the structural information of scene graph to provide a deeper understanding of surrounding environment.
The room type $\gamma_{t,i}$ is determined by querying the Vision-Language Model(VLM), which selects the most suitable option from a set of indoor room candidates based on the current image and surrounding objects. This result is then combined with the room types of nearby landmarks that have already been queried.The frontier property $\delta_{t,i}$ is a boolean variable obtained by querying the landmarks which point to the frontier in the landmark-instance mapping $\cD_t$.

\noindent\textbf{Agent-to-landmark prompts.}
The navigable message is encoded relative to the landmark $l_{t,now}$ where the agent is currently located. We calculate the distance and path from $l_{t,now}$ to $l_{t,i}$ on the landmark graph $\cL_t$ by means of Dijkstra Algorithm. The distance, combined with the direction, forms the navigable message of landmark $l_{t,i}$, \textit{e.g. as "Location: {Direction: Up, Path: $2\to0\to1\to3$, Distance: 3.2m}"}.
For navigation history, we set up a boolean variable explored property $\varepsilon_{t,i}$ representing whether the agent has explored this landmark before, avoiding repetitive exploration. 

\begin{figure*}[t]
\begin{center}
  \includegraphics[width=\linewidth]{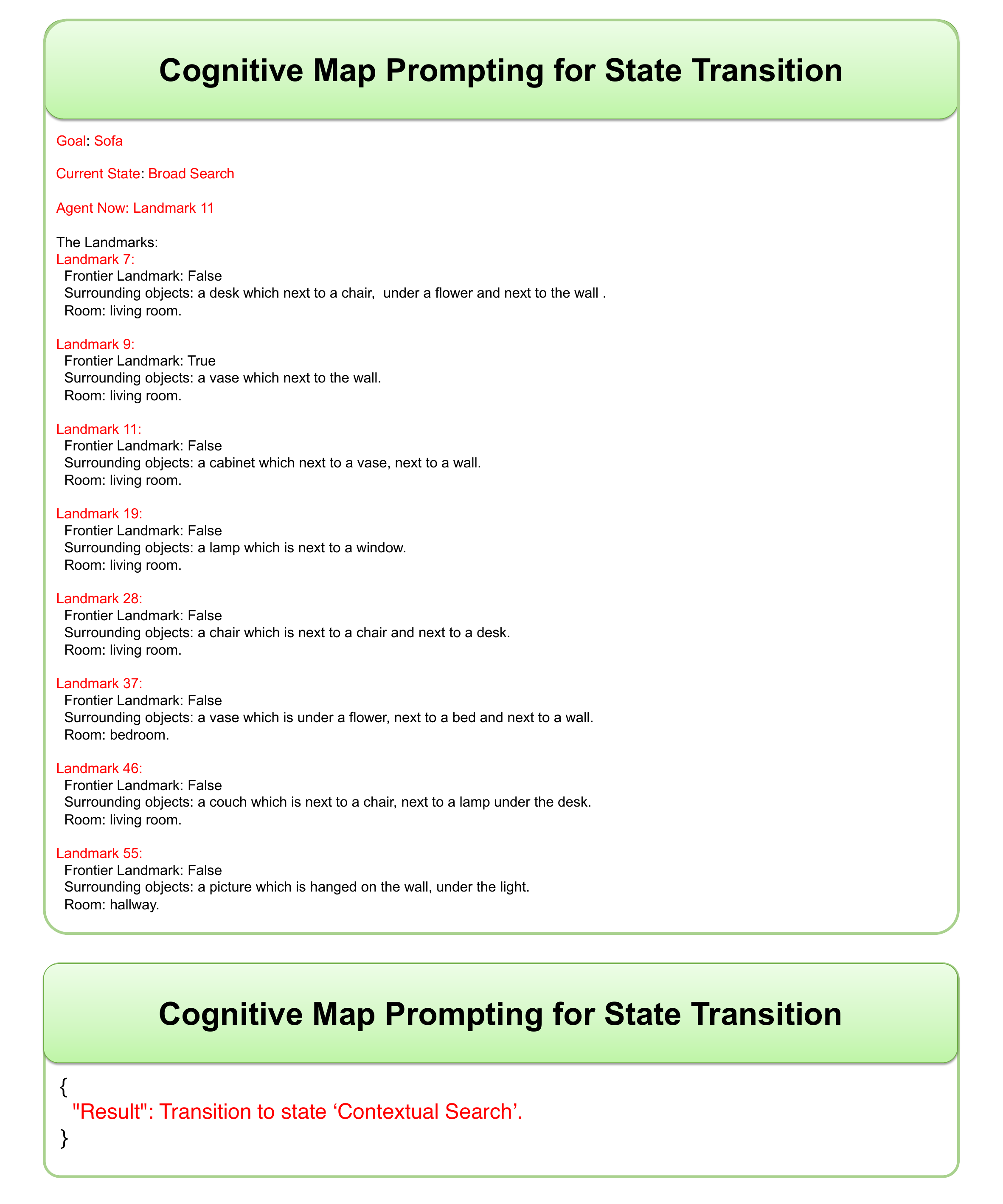}
\end{center}
\vspace{-10pt}
   \caption{\textbf{State Transition Prompting of CogNav and LLM Results.} We provide state transition prompting and state transition by LLM in an episode finding a sofa. The result by LLM is transited to state 'Contextual Search'.}
\label{fig:prompting_state}
\end{figure*}

\begin{figure*}[t]
\begin{center}
  \includegraphics[width=\linewidth]{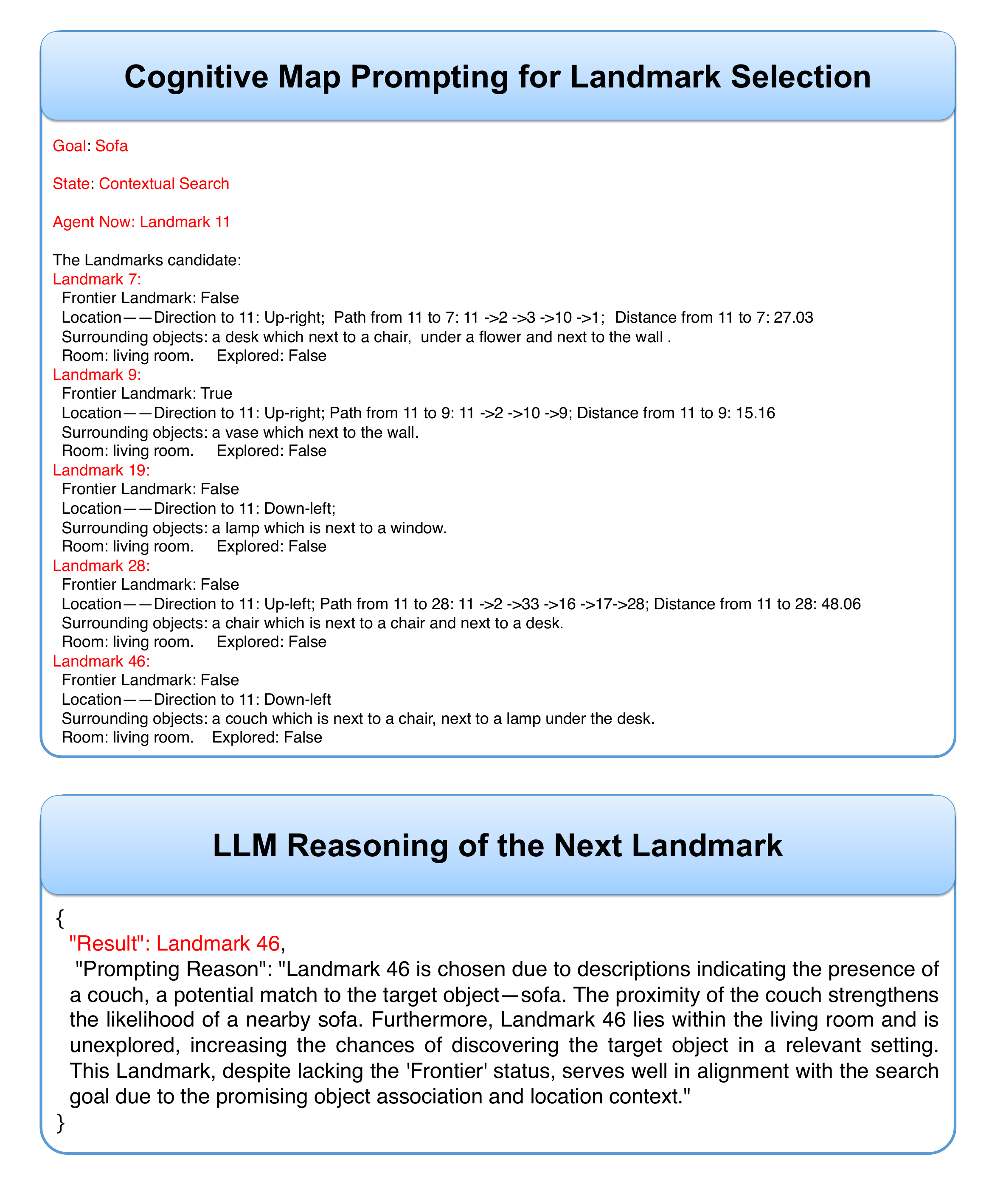}
\end{center}
\vspace{-10pt}
   \caption{\textbf{Cognitive Map Prompting of CogNav and LLM Results.} We provide landmark selection prompting and landmark selection by LLM in an episode finding a sofa. The current state is contextual search, we prompt the landmarks $\cL_{t,relevant}$ as candidates for next landmark selection.}
\label{fig:prompting_landmark}
\end{figure*}

The detail description of our state transition prompt and landmark selection prompt are shown in Figure.~\ref{fig:prompting_state} and Figure.~\ref{fig:prompting_landmark}, respectively.
\begin{table}[t]
\centering
\scalebox{0.9}{
\begin{tabular}{lcccc}
\hline
\multirow{2}{*}{\textbf{LLM}} & \multirow{2}{*}{\textbf{VLM}} & \multicolumn{3}{c}{\textbf{HM3D}}  \\ \cmidrule(r){3-5}
& &SR $\left ( \% \right ) \uparrow$ &SPL $\left ( \% \right ) \uparrow$  &DTG $\left ( m \right ) \downarrow$ \\
\hline
LLaMa3 & CogVLM2 &62.6 & 17.2 & 2.357\\
GPT-3 & CogVLM2 &64.1 &18.3 &1.882\\
LLaMA3 & GPT-4v &69.4 &22.2 &1.694\\
\hline
GPT-3 & GPT-4v & \textbf{72.5} & \textbf{26.2} & \textbf{1.255}\\ 
\hline
\end{tabular}
}
\caption{Comparison of various large language models on HM3D~\cite{ramakrishnan2021hm3d}. We replace GPT models \cite{brown2020language,achiam2023gpt} in CogNav with Llama3.1-8B-Instruct \cite{dubey2024llama} and Cogvlm2-llama3-chat-19B \cite{hong2024cogvlm2}.}
\vspace{-8pt}
\label{tab:llm}
\end{table}
\subsection{State-related landmarks in paper Sec.3.3.}
\begin{itemize}
    \item Broad Search (BS):  The landmark candidate in state BS are landmarks where the frontier property is \text{True}: $\cL_{t,frontier} = \left \{ l_{t,i} \in \cL_t \mid \delta _{t,i}=\text{True} \right \}$.
    \item Contextual Search (CS): The landmark candidate is changed to the landmarks which surrounding the relative instance node $n_r$ or the room type $\gamma_{r}$ which may contain the target object: $\cL_{t,relative} = \left \{ l_{t,i} \in \cL_t \mid \cD_t(l_{t,i})=n_r \vee \gamma_{t,i}=\gamma_{r}  \right \} $.
    \item Observe Target (OT): In state OT, landmarks in candidate are whose value of instance node which category is the target one $c$ in landmark-instance mapping $\cD_t$: $\cL_{t,target} = \left \{ l_{t,i} \in \cL_t \mid \cD_t(l_{t,i})=n_c  \right \} $.
    \item Candidate Verification (CV): When state CV is transformed, the landmark candidate selects landmarks in state OT however excludes landmarks explored before: $\cL_{t,verify} = \left \{ l_{t,i} \in \cL_t \mid \cD_t(l_{t,i})=n_c \wedge \varepsilon_{t,i} = \text{False} \right \} $.
    \item Target Confirmation (TC): While the instance $n_c$ is confirmed to be the target, the landmark closest to $n_c$ is chosen as the next term goal: $l_{t,target} = \arg\min_{l_{t,i} \in L_{t,target}} d(l_{t,i}, n_c)$, where $d(l_{t,i}, n_c)$ is the ground distance between the object $n_c$ and the landmark $l_{t,i}$. The navigation ends while the agent achieves this landmark.
\end{itemize}

\section{More Experiments}
\label{supsec:experiment}
\subsection{Results on Different LLMs of HM3D dataset}
We further replace the LLM and VLM to evaluate the reasoning and decision-making abilities of different large models in navigation. As shown in Table~\ref{tab:llm}, GPT (row 1) demonstrates superior analysis and decision-making capabilities in navigation tasks. Note that, after replacing it with two open-source large models (rows 1-3), our method still outperforms others on HM3D. 
\subsection{More Results of MP3D dataset}
\begin{figure}[t]
\centering
\begin{overpic}
[width=\linewidth]
{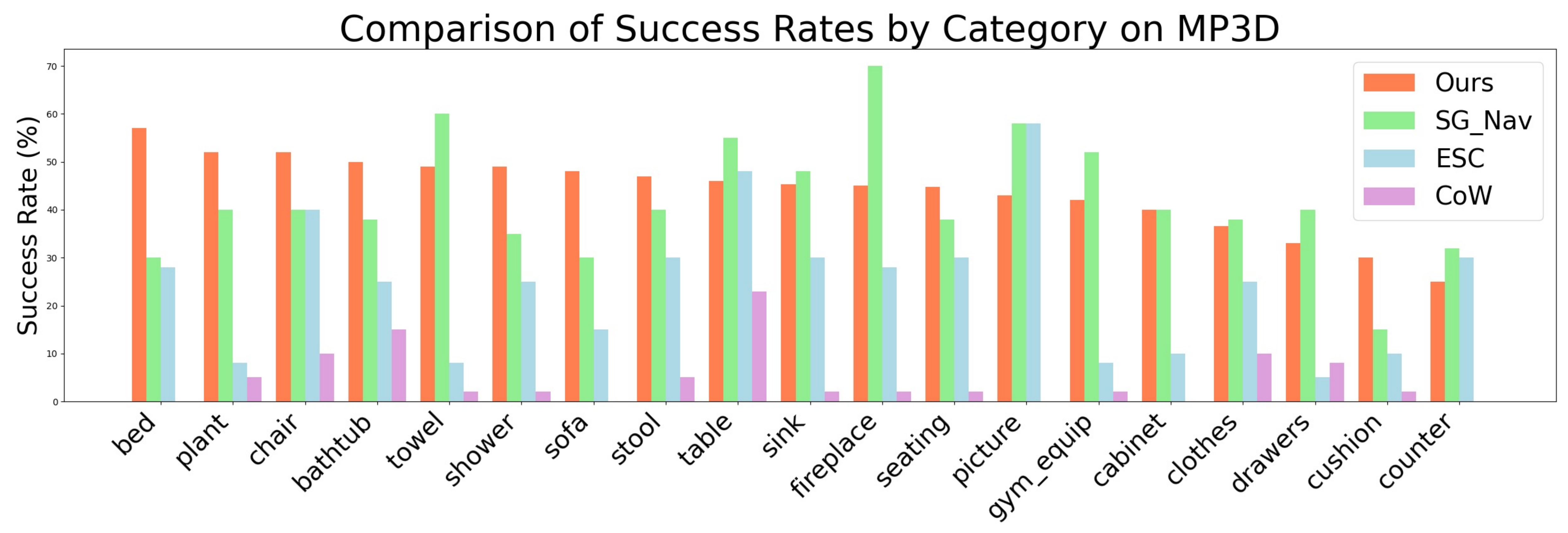}
\end{overpic}
\caption{The comparison of each category's success rate on MP3D~\cite{Matterport3D} with other three methods.
}
\label{fig:eachcategory}
\vspace{-8pt}
\end{figure}

We calculate the success rate (SR) of each category on the MP3D dataset \cite{Matterport3D} and compare with three methods Cow \cite{gadre2023cows}, ESC \cite{zhou2023esc} and SG\_Nav \cite{yin2024sg}. The results are demonstrated in Figure~\ref{fig:eachcategory}. Our success rate of category \texttt{fireplace} and \texttt{picture} are lower than SG-Nav \cite{yin2024sg} because of the unrecognizable of these two categories in some viewpoints. In other categories, our framework have the highest success rate, especially in category \texttt{bed}, \texttt{plant}, \texttt{bathtub} and \texttt{sofa}, our success rate metric is far superior to other three methods.
More qualitative results can be fount in the first lines of Figure.~\ref{fig:gallery2}

\renewcommand{\figurename}{Fig}
\begin{figure}[t]
\begin{center}
  \includegraphics[width=\linewidth]{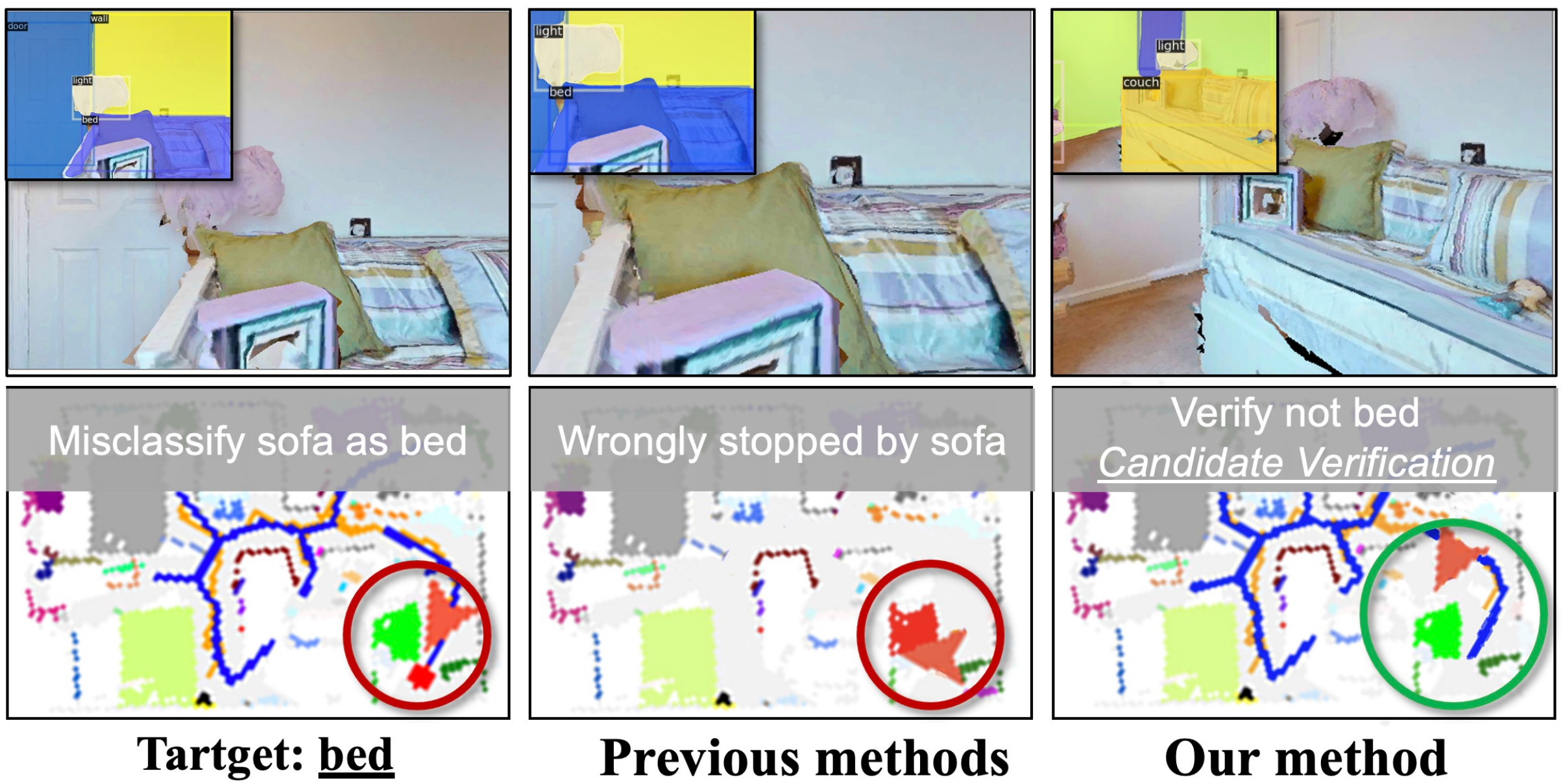}\vspace{-10pt}
\end{center}
   \caption{Target sofa is misclassified to bed (Col.1), previous methods totally believe the result and stop when approach (Col.2). While, our method verify it in another view and correct the error (Col.3).}
\label{fig:example}
\vspace{-7pt}
\end{figure}

We also provide an example of navigational reasoning in Fig.~\ref{fig:example}, where our method demonstrates error correction capability (with candidate verification) compared to previous approaches.

For fail case, we analyze in HM3D (Val) with  277/1008, among which
$35.7\%$ are due to persistent failures in object detection, $64.3\%$ because of target inaccessibility (\textit{e.g.} mesh artifacts, stairs). 

\subsection{Time Analysis of Our Framework}
We measured the time overhead of different components in our framework, calculating the average time per execution step and the average frequency per episode for each component. The results, summarized in Table \ref{tab:time}, represent averages across all episodes on HM3D.
Note that our time overhead computation includes failure cases where the step limit of 500 is reached, leading to an increased number of LLM queries. While the GPT-4V query process incurs a higher time overhead per query, its lower frequency minimizes its overall impact. The scene graph updating process accounts for the majority of the time overhead, primarily due to the high frequency of segmentation and detection operations performed by OpenSEED~\cite{zhang2023simple}.
Despite these factors, our framework ensures effective real-time navigation in unknown environments.


\begin{table}[t]
\centering
\scalebox{0.9}{
\setlength{\tabcolsep}{1mm}{
\begin{tabular}{lcc}
\hline
\textbf{Part of Framework} & \textbf{Time (s)} & \textbf{Frequency} \\ \hline
Scene graph Updating & 1.8 & 230\\
Landmark Graph Updating & 0.85 & 26 \\
Room-type Query with GPT-4v & 1.8 & 23 \\
Cognitive Map Prompting & 0.6 & 15 \\
Cognitive Process Modeling with GPT-4v & 3.4 & 15 \\
\hline

\end{tabular}
}
}
\caption{The average time and frequency cost in each part of all episodes on HM3D and MP3D, which includes the time overhead of the fail cases running the limited time budget of 500 steps. Time and frequency means average time for each part and average frequency in one episode, respectively.}
\vspace{-12pt}
\label{tab:time}
\end{table}
\subsection{Real-World Experiment Details}
\noindent\textbf{Real-World Setup.}
To deploy CogNav in the real world, we build a custom robot comprising an automated guided vehicle, robotic arm, an RGB-D camera and a master control computer. For the automated guided vehicle, we utilized Water II~\cite{water_two}, which features a single-wire lidar for implementing indoor navigation algorithms and localization. The robotic arm selected is the E05~\cite{E05}, providing flexibility and convenience for adjusting camera poses. The RGB-D camera used is the Microsoft Azure Kinect DK~\cite{azure_kinect}. The master control computer are equipped with an Intel$^\circledR$ Core$^\textbf{TM}$ i7-10700K @ 2.9GHz $\times$ 8. We run the memory building phase on the control workstation with an Nvidia GeForce RTX 3090Ti GPU with 24GB of memory, paired with an Intel$^\circledR$ Core$^\textbf{TM}$ i9-12900K @ 3.9GHz $\times$ 16 and 32GB of RAM. The system operates on ROS Noetic Ninjemys as the software platform. In an object navigation episode, the master control computer transmits the RGB-D data stream and the 2D pose from the automated guided vehicle to the control workstation via ROS topics. The workstation processes this information as input for CogNav. Upon receiving the target landmark's 2D pose from CogNav, the workstation directs the robot to move to the specified location. These operations repeat iteratively until the target object is successfully located. We also validate our pipeline on a Unitree quadruped, the navigation processes are demonstrated in the demo video.

\begin{figure*}[t]
\begin{center}
  \includegraphics[width=1.05 \linewidth]{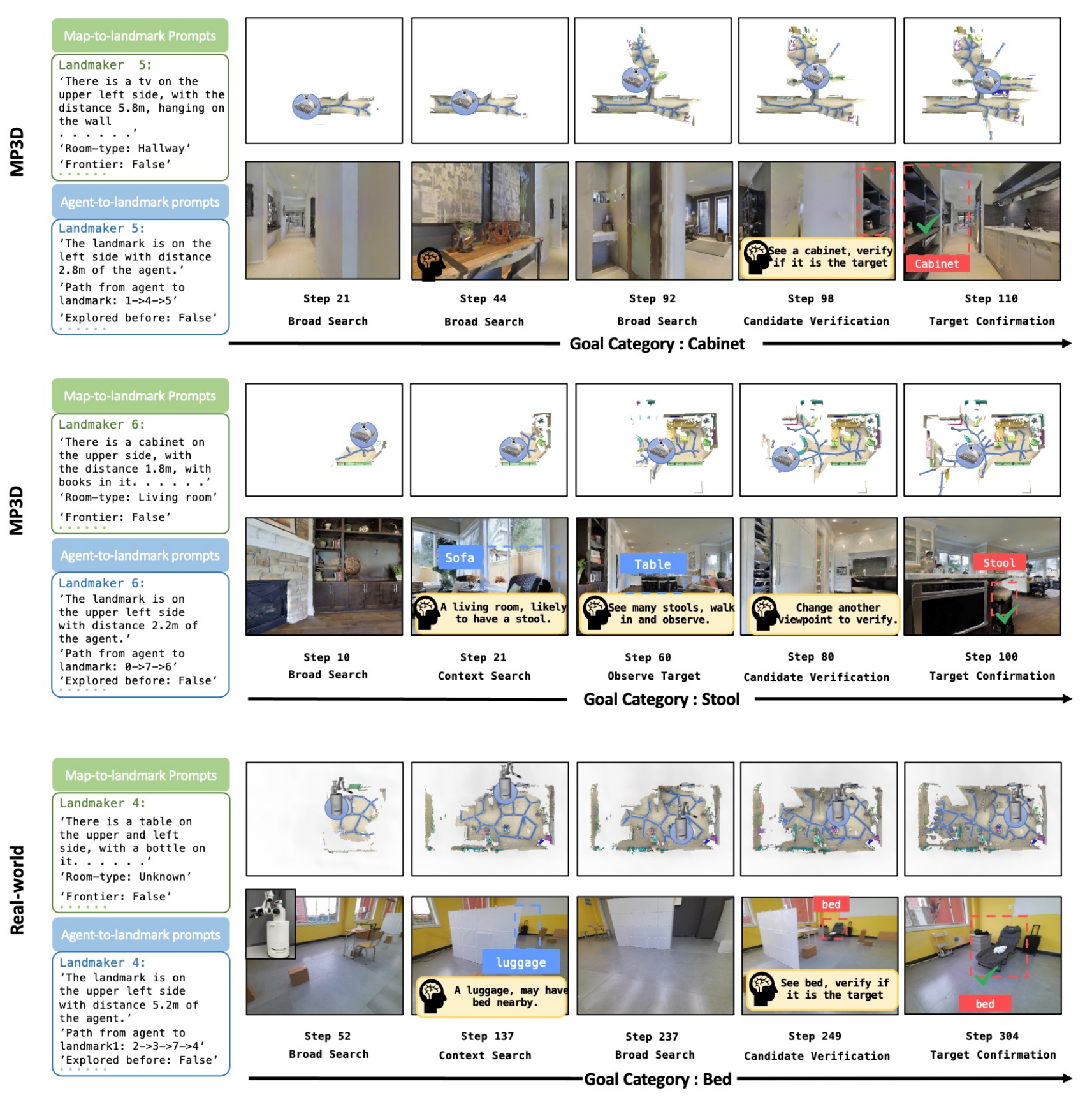}
\end{center}
\vspace{-10pt}
   \caption{\textbf{Navigation process visualization of CogNav.} We provide visual results of navigation process for one synthetic scene and one real-world one. Cognitive maps encode scene information and facilitate landmark prompting, enabling the agent to explore environments efficiently and identify target objects accurately.}
\label{fig:gallery2}
\end{figure*}

\noindent\textbf{Qualitative Results.} 
To simulate a home environment, we organized a room of approximately $60m^{2}$ into distinct functional areas, including a work area, a rest area, and a dining area, separated by partitions. To validate the feasibility of CogNav on robots, we carefully selected target objects of varying sizes, categories, and properties within the scene. Additional qualitative results from the real-world experiments are shown in the last line of Figure~\ref{fig:gallery2}.

\begin{figure*}[t]
\begin{center}
  \includegraphics[width=\linewidth]{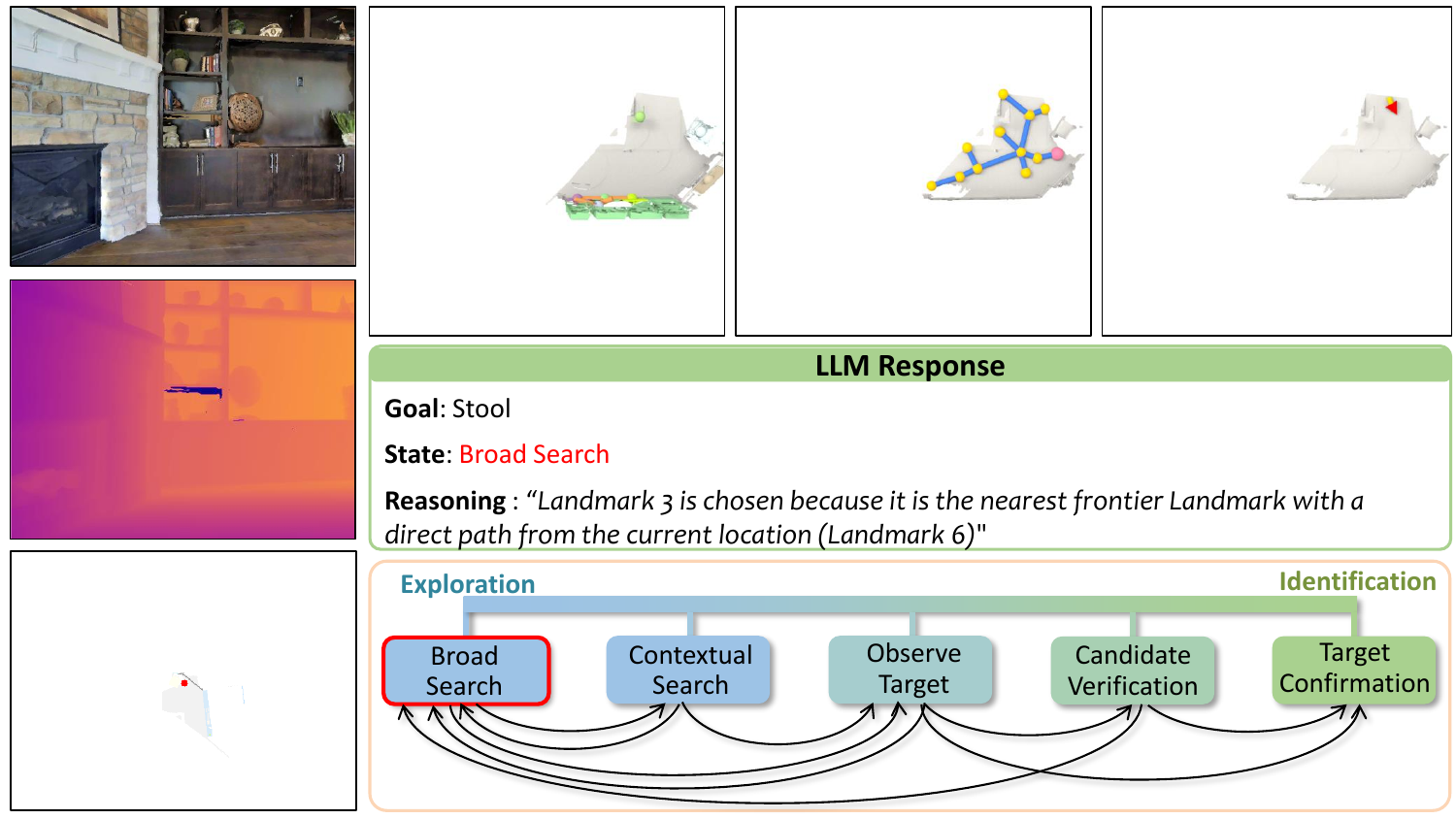}
  1
\end{center}
\begin{center}
  \includegraphics[width=\linewidth]{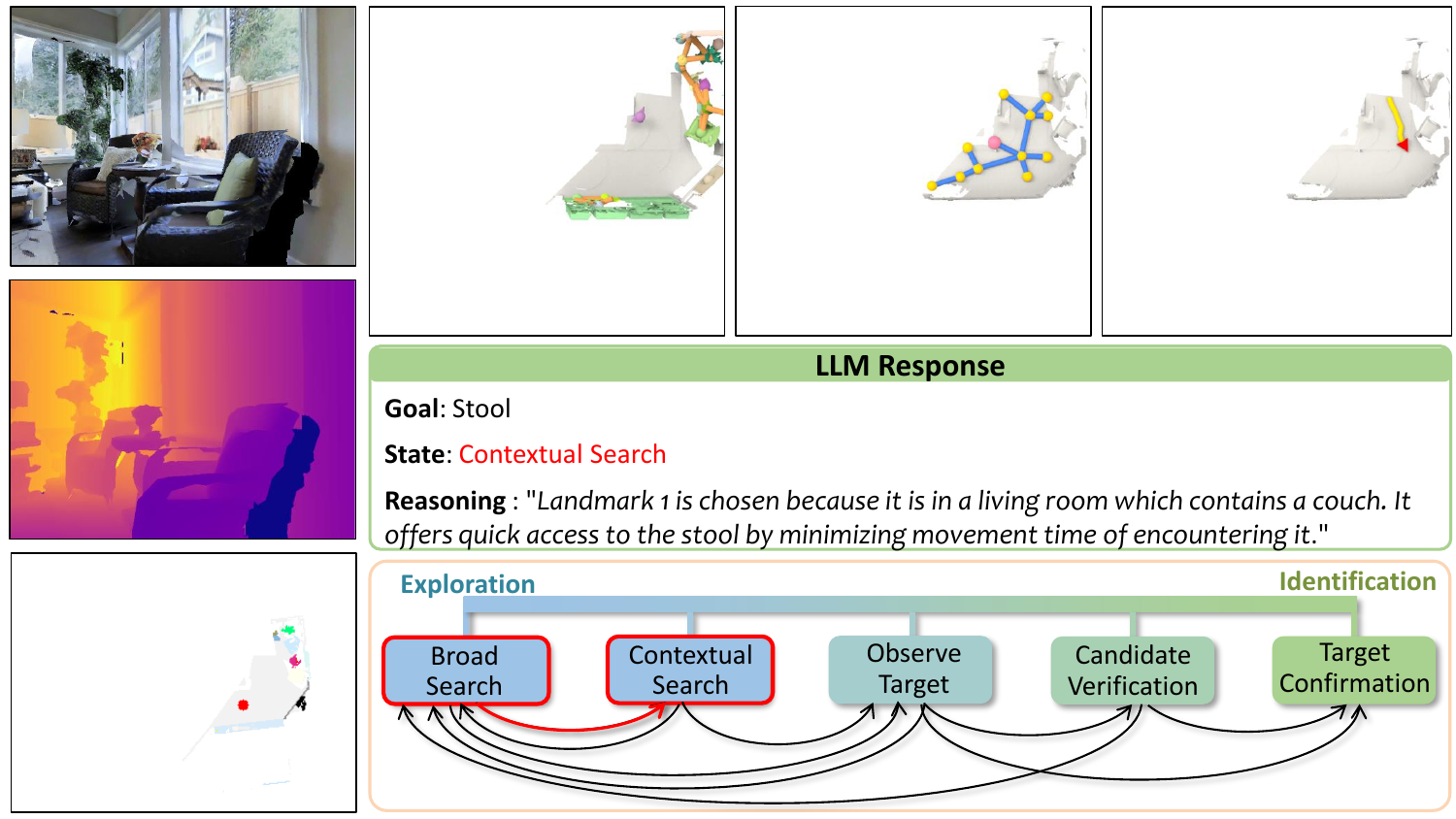}
  2
\end{center}
    \caption{\textbf{State transition in navigation process of finding a stool.}}
\label{fig:navigation_stool}
\end{figure*}

\begin{figure*}[t]
\begin{center}
  \includegraphics[width=\linewidth]{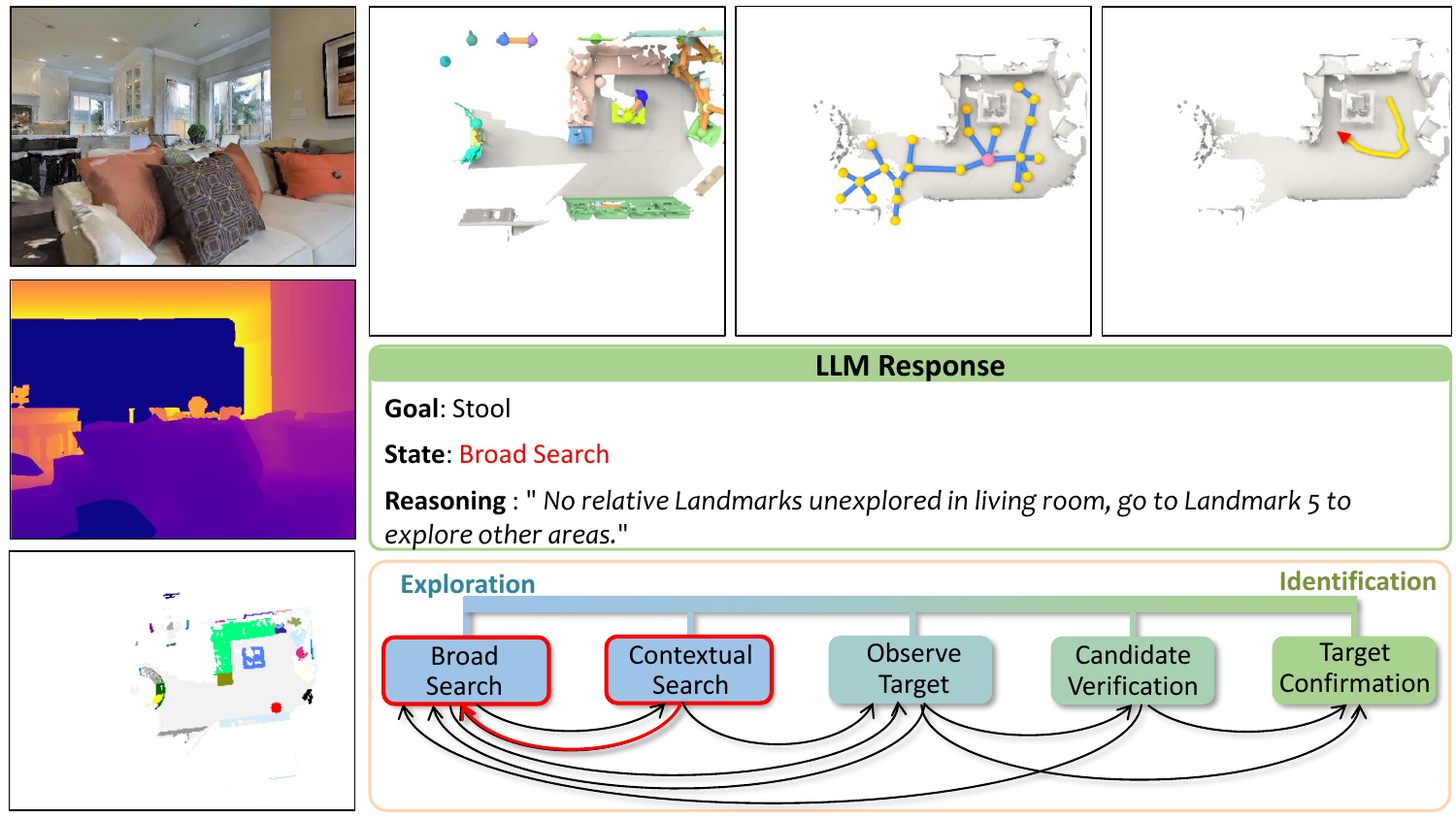}
  3
\end{center}
\begin{center}
  \includegraphics[width=\linewidth]{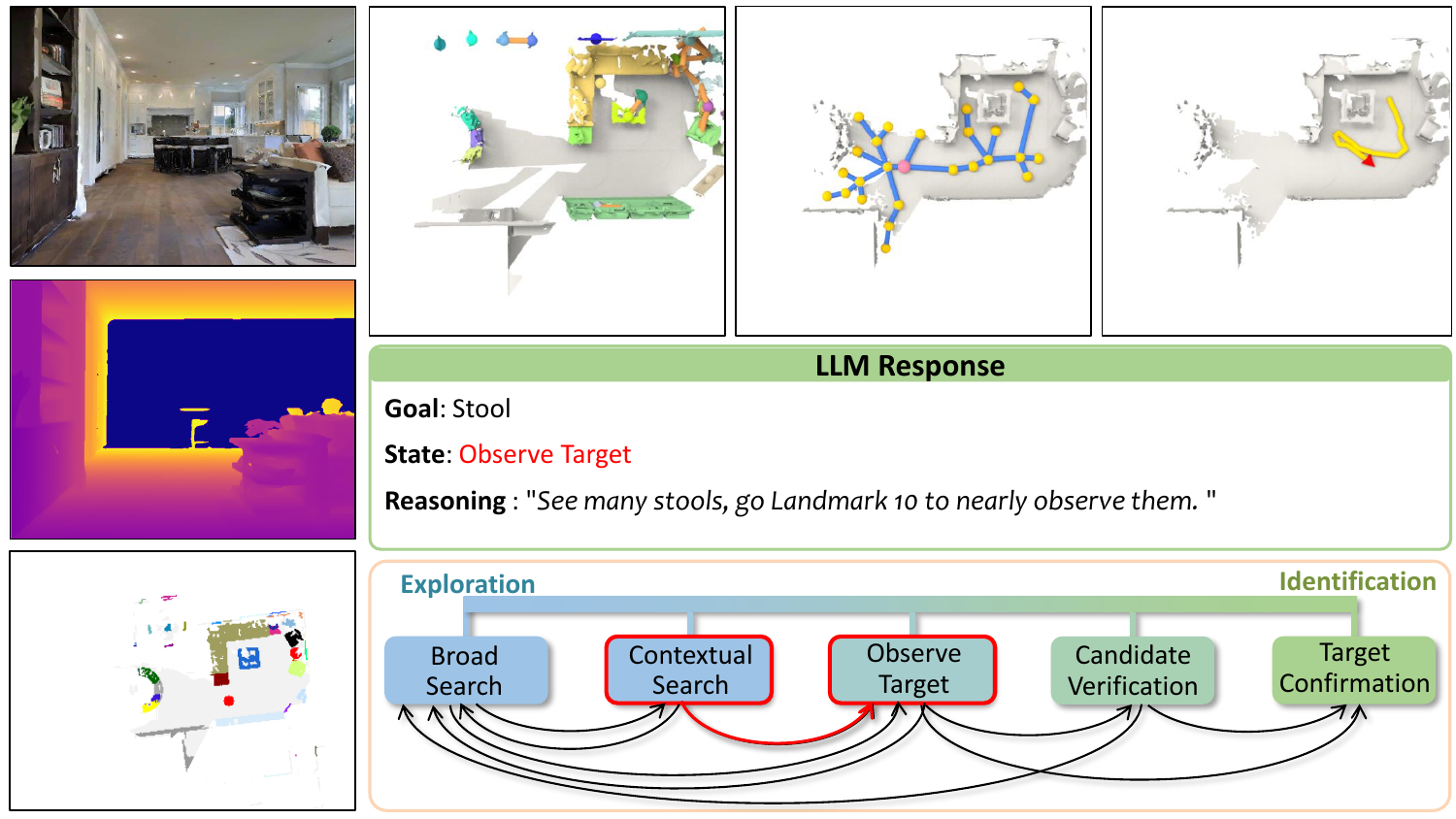}
  4
\end{center}
    \caption{\textbf{State transition in navigation process of finding a stool.}}
\label{fig:navigation_stool2}
\end{figure*}
\begin{figure*}[t]
\begin{center}
  \includegraphics[width=\linewidth]{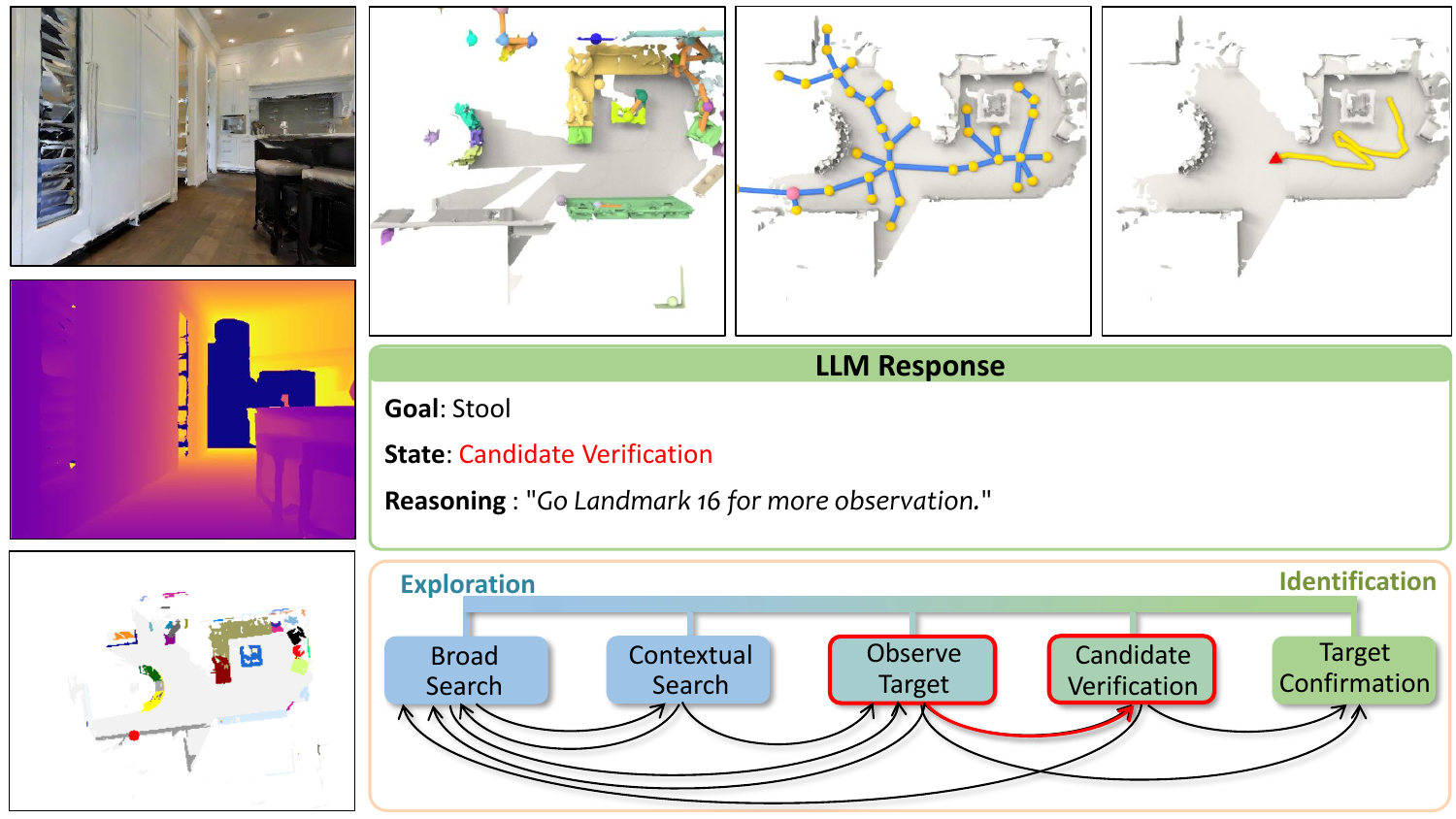}
  5
\end{center}
\begin{center}
  \includegraphics[width=\linewidth]{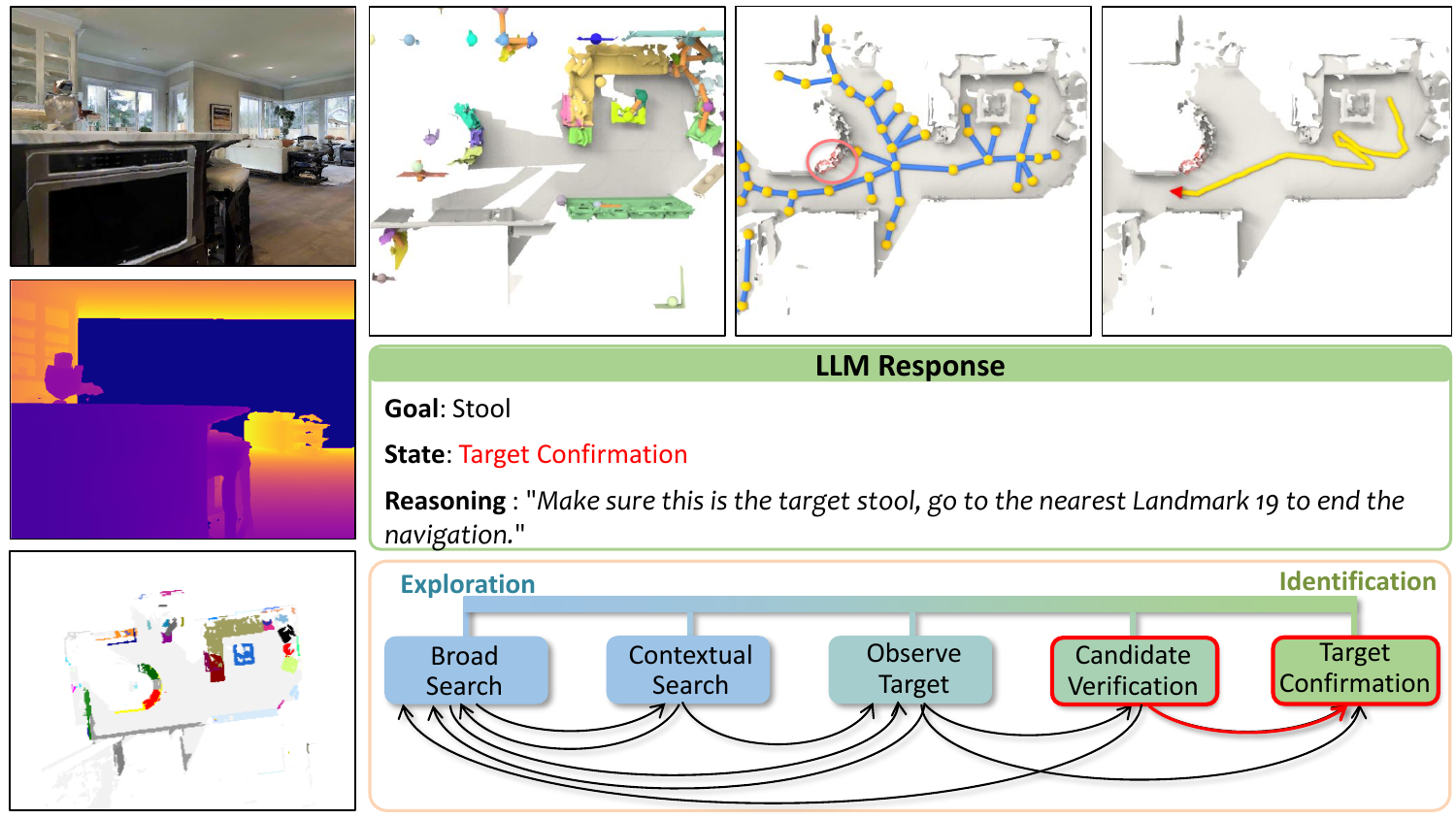}
  6
\end{center}
    \caption{\textbf{State transition in navigation process of finding a stool.}}
\label{fig:navigation_stool3}
\end{figure*}
\clearpage
{
    \small
    \bibliographystyle{ieeenat_fullname}
    \bibliography{main}
}
